\pdfoutput=1
\documentclass[11pt]{article}
\usepackage{multirow}
\usepackage[normalem]{ulem}
\usepackage{tabularray}
\usepackage{color}
\usepackage{makecell}
\usepackage[table]{xcolor} % 如果要着色，需要引入 xcolor 包
\usepackage{booktabs}      % 更美观的表格线
\usepackage[]{acl}
\usepackage{tcolorbox}
\usepackage{arydshln}
\usepackage{times}
\usepackage{latexsym}
\usepackage{array} % required for text wrapping in tables
\usepackage{framed}
\usepackage{verbatim}
\usepackage{amsmath}
\usepackage{amssymb}
\usepackage{cuted}
\usepackage{xcolor}
\definecolor{DarkGreen}{RGB}{0,100,0} 
\definecolor{DarkRed}{HTML}{960018}
\usepackage[T1]{fontenc}
\usepackage[utf8]{inputenc}
\usepackage{inconsolata}
\usepackage{microtype}
\usepackage{graphicx}
\usepackage{array}

\usepackage{soul}

\title{ReSURE: Regularizing Supervision Unreliability for \\Multi-turn Dialogue Fine-tuning}

\author{Yiming Du$^{1}$, Yifan Xiang$^{1}$, Bin Liang$^{1}$, \textbf{Dahua Lin}$^{1}$, \\ \textbf{Kam-Fai Wong}$^{1}$\thanks{\; Corresponding authors}, \textbf{Fei Tan}$^{2}$\footnotemark[1]\\
  $^1$The Chinese University of Hong Kong 
  $^2$ East China Normal University \\
    \texttt{\{ydu, kfwong\}@se.cuhk.edu.hk},  \texttt{ftan@mail.ecnu.edu.cn} }
\begin{document}
\maketitle
\begin{abstract}
Fine-tuning multi-turn dialogue systems requires high-quality supervision but often suffers from degraded performance when exposed to low-quality data. Supervision errors in early turns can propagate across subsequent turns, undermining coherence and response quality. Existing methods typically address data quality via static prefiltering, which decouples quality control from training and fails to mitigate turn-level error propagation. In this context, we propose \textbf{ReSURE} (Regularizing Supervision UnREliability), an adaptive learning method that dynamically down-weights unreliable supervision without explicit filtering. ReSURE estimates per-turn loss distributions using Welford’s online statistics and reweights sample losses on the fly accordingly. Experiments on both single-source and mixed-quality datasets show improved stability and response quality. Notably, ReSURE enjoys positive Spearman correlations (0.21 $\sim$ 1.0 across multiple benchmarks) between 
response scores and number of samples regardless of data quality, which potentially paves the way for utilizing large-scale data effectively. Code is publicly available at \url{https://github.com/Elvin-Yiming-Du/ReSURE_Multi_Turn_Training}.

%It exhibits consistent optimization trends across benchmarks, with Spearman correlations of 0.21 (In-domain Test), 1.00 (MT-Bench), and 0.80 (MT-Bench-Ext), even under increasing supervision noise.

%ReSURE offers a lightweight yet effective solution for robust multi-turn dialogue fine-tuning.
% This design enables stable optimization across multi-turn dialogues by preserving contextual coherence and mitigating the compounding effect of early-turn errors.
\end{abstract}

% \begin{figure}[!t]
% \centering
% \includegraphics[width=\linewidth]{imgs/intro.pdf}
% \caption{Intro.}
% \label{method}
% \end{figure}
\begin{figure*}[t!]
  \centering
  \includegraphics[width=\linewidth]{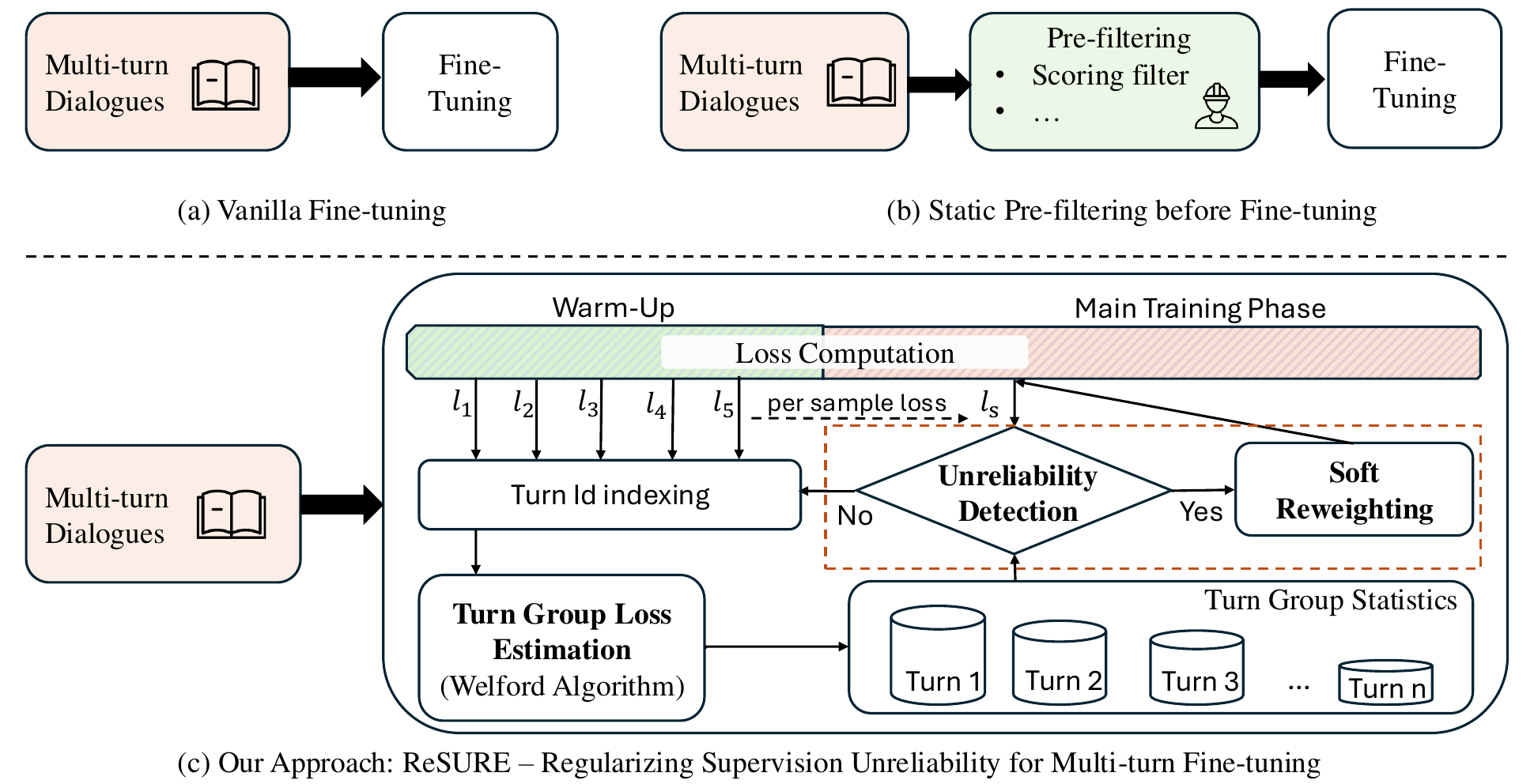}
  \caption{Overview of Training Paradigms: Traditional Fine-tuning, Pre-filtering, and ReSURE.}
  \label{f-main_framework}
\end{figure*}

\section{Introduction}
Multi-turn dialogue systems are fundamental to both task-oriented \cite{xu-etal-2024-rethinking} and open-domain conversational agents \cite{lu2023memochat, sun-etal-2024-parrot}, enabling coherent and natural interactions. However, fine-tuning remains challenging due to reliance on large-scale multi-turn datasets \cite{ChatAlpaca, zhao2024wildchat, OpenAssistant} that mix human and synthetic data of varying quality \cite{openai2023gpt4, zhan2025mathsmith}. In such settings, supervision errors in early turns often propagate across later ones, compounding inconsistencies and degrading coherence \citep{hu2025steering, yi2024survey}. This issue is further exacerbated by mismatches between training supervision and evaluation criteria, making it difficult for models to recover from early-turn noise or learn turn-consistent behavior \cite{zheng2023judging, kwan2024mt, wu2023self, chen2023alpagasus, li-etal-2024-quantity, zhou2024lima}. As datasets scale, conventional fine-tuning approaches assume uniformly reliable supervision and struggle to distinguish between clean and noisy signals, often overfitting to noise or discarding useful samples \citep{hase-etal-2024-unreasonable}.

A common strategy to mitigate noisy supervision is static pre-filtering before fine-tuning \cite{wang2024survey}, aiming to remove low-quality or incomplete samples. However, such heuristic methods \cite{cao2023instruction} overlook the hierarchical nature of multi-turn dialogues, leading to over-filtering and loss of informative turns. Other approaches enhance robustness by injecting synthetic noise \cite{wu2022noisytune,gu-etal-2024-cmr}, but often lack principled mechanisms to regulate supervision quality during training.

To address these limitations, we propose \textbf{ReSURE} (Regularizing Supervision UnREliability), an adaptive fine-tuning framework that dynamically adjusts loss contributions from unreliable supervision signals. We define such supervision as samples that consistently yield high or unstable losses during training \citep{wang2024absinstruct,zhang-etal-2024-balancing}. Observing that later turns are more susceptible to supervision noise due to increased contextual complexity \citep{zheng2023judging, kwan2024mt}, ReSURE groups samples by turn depth and tracks per-group loss statistics via Welford’s algorithm \citep{welford1962method, efanov2021welford}. Samples with abnormally high losses are softly reweighted to reduce instability while preserving gradient signal. This turn-aware design ensures that difficult turns are not over-penalized and early-turn errors do not dominate optimization, thereby stabilizing training and enhancing contextual coherence in multi-turn dialogue.

Experimental results show that ReSURE enables consistent optimization across multi-turn benchmarks, including MT-Bench, MT-Bench-Ext, and In-Domain-Test. Under mixed datasets with progressively added noisy or off-distribution samples, ReSURE consistently maintains or improves performance, achieving positive Spearman correlations (0.21, 1.00, 0.80), while Vicuna-Tuning \citep{vicuna2023} shows degradation and other baselines fluctuate. To simulate task-level noise, we incorporate GSM8K \citep{cobbe2021training} and find that ReSURE preserves generalization. These findings highlight ReSURE’s robustness to both supervision noise and task drift. Unlike static filtering methods such as DeBERTa-based data selection \citep{he2020deberta, he2021debertav3}, ReSURE achieves these gains without manual intervention. Moreover, combining ReSURE with pre-filtering yields further improvements, indicating their complementarity.

Our key contributions are as follows:
\begin{itemize}
    \item We propose \textbf{ReSURE}, a turn-aware fine-tuning framework that preserves positive optimization in multi-turn instruction tuning under unreliable supervision.
    \item ReSURE avoids manual data filtering and seamlessly integrates with instruction-tuning pipelines, supporting robustness under both supervision and task-level noise.
    \item Extensive experiments across in-domain, MT-Bench, and MT-Bench-Ext show consistent gains and positive optimization trends (Spearman: 0.21, 1.00, 0.80), with further improvement when combined with static filtering.
\end{itemize}
\section{Related Work}

\subsection{Multi-turn Dialogue Fine-tuning}
%Advances in LLM fine-tuning \citep{hu-etal-2023-llm,hu2021lora,dettmers2024qlora} significantly improve performance and develop a family of models \citep{liu2024dora,zhao2024galore,meng2024pissa}} 

Recent advances in LLM fine-tuning \citep{hu-etal-2023-llm,hu2021lora,dettmers2024qlora} have enabled strong performance on single-turn tasks \citep{liu2024dora,zhao2024galore,meng2024pissa}, but multi-turn dialogue remains challenging. Prior work addresses this via optimization techniques like reinforcement learning \citep{zhang-etal-2025-dadpo} and preference modeling \citep{sun-etal-2024-parrot,shani2024multi}, or through data augmentation and inductive construction \citep{maheshwary2024m2lingual,ou-etal-2024-inductive, du2025bridging}. However, these lines remain disconnected, and challenges like data curation cost, weak generalization, and inconsistent supervision persist. Our work bridges this gap by jointly addressing turn-level supervision and data noise in a unified framework.

\subsection{Data Selection in LLM Finetuning}

Although the scale of data is crucial in LLM fine-tuning, selecting fewer high-quality data points can lead to better performance than using the entire dataset \citep{wu2023self,chen2023alpagasus}, highlighting the significance of data selection. In terms of data quality assessment \citep{wang2024survey}, data selection methods can be grouped into three categories: (1) GPT-based scoring, which relies on prompting ChatGPT with predefined rubrics \citep{chen2023alpagasus,lu2023instag,xu2023variety,liu2024what,du2023mods}; (2) model-based scoring, where an LLM is trained to evaluate instances under a learned policy \citep{li2023self,li-etal-2024-one,anonymous2024selfevolved}; and (3) indicator-based methods \citep{lu-etal-2023-makes}, which estimate data quality via inference loss \citep{cao2023instruction,wang-etal-2024-reward} or handcrafted conversation metrics \citep{wei2023instructiongpt}.

Although these works emphasize the importance of data selection, they often produce uninterpretable results, suffer from limited applicability and randomness, and demand prohibitively high training costs. These limitations lead to low feasibility in both training and generalization as models evolve. In addition, prior approaches perform data selection independently of the training process, failing to capture and leverage end-to-end feedback during training, which is a key focus of our work.

\section{Methodology}
%We present ReSURE, an adaptive fine-tuning mechanism that dynamically reweights per-sample losses based on their reliability. 
By monitoring loss statistics in a turn-aware manner using Welford’s online algorithm, ReSURE identifies unstable supervision signals and adjusts their training influence without explicit filtering. This design stabilizes optimization and preserves coherence in multi-turn dialogue settings.

Specifically, in multi-turn fine-tuning, each training sample consists of a dialogue with multiple user–assistant turns. The model is trained to minimize the cross-entropy loss over the supervised tokens. ReSURE modifies this objective by introducing a dynamic weight $w_s$ for each sample:

\begin{equation}
\mathcal{L}_{\text{ReSURE}} = \frac{1}{S} \sum_{s=1}^S w_s \cdot \ell_s,
\end{equation}
where $\ell_s$ denotes the loss for sample $s$, $S$ denotes the number of samples in the mini-batch, and $w_s$ is computed based on turn-aware loss statistics (see Sec.~\ref{subsec:soft_reweighting}).

\subsection{Turn Group Loss Estimation}

Supervised fine-tuning in multi-turn dialogue is complicated by uneven supervision quality across dialogue depths. Early turns are typically short, contextually grounded, and easier to align with reference responses. In contrast, later turns often involve complex phenomena such as context accumulation, topic shifts, and implicit reasoning, which increase supervision noise and model uncertainty \citep{zheng2023judging, kwan2024mt}.

To address this, ReSURE groups training samples by their maximum supervised turn group index $b \in \{1, \dots, N\}$, where $N$ denotes the maximum number of turns per dialogue. For each $b$, we maintain turn-specific online loss statistics—namely, a running mean $\mu_s^{(b)}$ and standard deviation $\sigma_s^{(b)}$ of the per-sample loss—computed using Welford’s algorithm:
\begin{equation}
\mu_s^{(b)} = \mu_{t-1}^{(b)} + \frac{l_s - \mu_{s-1}^{(b)}}{t^{(b)}} ,
\label{eq:mean}
\end{equation}
\begin{equation}
SSD_s^{(b)} = SSD_{t-1}^{(b)} + (l_s - \mu_{t-1}^{(b)})(l_s - \mu_s^{(b)}),
\label{eq:m2}
\end{equation}
\begin{equation}
\sigma_s^{(b)} = \sqrt{\frac{M2_s^{(b)}}{t^{(b)} - 1}}.
\label{eq:variance}
\end{equation}

Here, $SSD_s^{(b)}$ denotes the Sum of Squared Deviations from the current mean $\mu_s^{(b)}$, used to compute the variance, and $s^{(b)}$ is the number of samples assigned to group $b$ up to sample $s$. Only samples within each group $b$ contribute to its own loss statistics, enabling turn-aware normalization. All statistics are initialized to zero and updated only upon observing the first reliable sample in each turn group. This design avoids over-penalizing high-turn samples that are harder, while ensuring stable optimization on easier low-turn cases. By aligning loss treatment with dialogue structure, it provides an inductive bias that helps the model calibrate supervision trust by turn depth.

\subsection{Unreliability Detection}
After warm-up, ReSURE detects unreliable supervision by identifying loss outliers with respect to turn-specific distributions. For each dialogue turn, we maintain the running mean $\mu^{(b)}$ and standard deviation $\sigma^{(b)}$ of per-sample loss using Welford’s algorithm. A sample is flagged as unreliable if its loss $l_s$ exceeds the threshold:

\begin{equation}
\tau_s^{(b)} = \mu_s^{(b)} + \alpha \cdot \sigma_s^{(b)},
\label{eq:threshold}
\end{equation}
where $\alpha$ is a fixed anomaly factor. While classical outlier detection often adopts $\alpha$ under Gaussian assumptions, we use $\alpha = 1.0$ to increase sensitivity to moderate deviations, following practices in robust training and loss-based re-weighting \citep{zhang2020generalized}.

If a sample is identified as unreliable ($l_s > \tau_s^{(b)}$), its loss is downweighted using soft reweighting (see Sec.~\ref{subsec:soft_reweighting}) but excluded from the update of running statistics, and the statistics $\mu^{(b)}$ and $\sigma^{(b)}$ remain unchanged. In contrast, if $l_s \leq \tau_s^{(b)}$, the sample is treated as reliable and its loss is incorporated into the Welford updates as defined in Eqs.~\eqref{eq:mean},~\eqref{eq:m2}, and~\eqref{eq:variance}. This conditional update mechanism ensures that the estimated statistics remain stable in the presence of supervision noise while still adapting to distributional shifts in reliable examples.

\subsection{Soft Reweighting}
\label{subsec:soft_reweighting}

Rather than discarding high-loss samples, ReSURE applies a soft reweighting strategy to reduce their influence while retaining informative gradients. For samples with $l_s > \tau_s^{(b)}$, the adjusted loss is computed using a decayed weight:

\begin{equation}
w_s = \max\left(\epsilon_s, \exp\left(-\frac{l_s - \tau_s^{(b)}}{\tau_s^{(b)}}\right)\right),
\end{equation}
\begin{equation}
\tilde{l}_s = w_s \cdot l_s,
\end{equation}
where $\tau_s^{(b)}$ is the turn-specific loss threshold and $\epsilon_s$ denotes a dynamic floor, computed as the 5th percentile of the current batch’s weight distribution. This adaptive lower bound ensures that even high-loss samples retain a minimal contribution, preventing vanishing gradients while adapting to overall batch variability.

Unlike fixed heuristics, this percentile-based formulation provides a data-driven way to preserve training signal from difficult or ambiguous cases. It aligns with findings in robust optimization that emphasize the importance of soft suppression rather than hard filtering for handling uncertain supervision \citep{ren2018learning, zhang2020generalized}. The hyperparameters follow common practices in robust optimization and noise-robust learning. We set $\alpha = 1.0$ to correspond to a one-standard-deviation threshold, a setting widely used in variance-based outlier detection and loss reweighting \citep{ren2018learning}. This value strikes a balance between sensitivity to moderate deviations and stability under noise. For $\epsilon$, we use the 5th percentile of the current batch’s weight distribution, following percentile-based reweighting strategies shown to be effective in preserving informative gradients while suppressing extreme outliers \citep{lei2025mining}. Empirical tuning on validation sets confirmed that more aggressive cutoffs (e.g., 1st percentile) over-penalize valid hard examples, while more moderate ones (e.g., 10th or 25th) reduce noise suppression efficiency.

\section{Experiments}

\subsection{Evaluation on Datasets}
\label{Evalation on Datasets}

There are multiple open-source and high-quality multi-turn dialogue datasets, which are generated by both humans and LLMs. Table \ref{tab:Datasets} in section \ref{sec:datasets intro} of Appendix presents the datasets used in this work and their features, including ShareGPT \citep{ShareGPT}, WildChat \citep{zhao2024wildchat}, OpenAssistant \citep{kopf2024openassistant}, ChatAlpaca \citep{ChatAlpaca}, M2Lingual \citep{maheshwary2024m2lingual}, and UltraChat \citep{ding2023enhancing}. Motivated by benchmarks on LLM evaluation \citep{zheng2023judging,kwan-etal-2024-mt} and MoDS \cite{du2023mods}, we evaluate these datasets by \textbf{GPT} and \textbf{reward model}, respectively. This dual-evaluation strategy offers complementary insights and enables a comprehensive evaluation on dataset quality.\par
\textbf{Evaluation by GPT.} Recent benchmarks on LLM evaluation \citep{zheng2023judging,kwan-etal-2024-mt,radziwill2017evaluating} emphasize relevance, helpfulness, and accuracy, while also acknowledging ethical considerations. Besides, prior work on human dialogue \citep{10.1016/j.csl.2015.11.001} highlights the importance of information density. Thus, we propose a benchmark evaluating conversations in four independent aspects: \textit{Connection}, \textit{Quality}, \textit{Information Density} and \textit{Friendliness}.\par
The evaluation is carried out using GPT-4o, which is widely adopted in evaluation works \citep{zheng2023judging,kwan-etal-2024-mt,bai-etal-2024-mt}. The designs of criteria, prompts and data pre-processing are detailed in section \ref{sec:evaluation prompts} of Appendix. For the evaluation on each aspect, one hundred conversations are sampled independently and randomly, and the evaluation on each conversation of each dataset is also independent. The score of each aspect of a dataset is defined as the average score of the sampled conversations in this aspect.\par
\textbf{Evaluation by reward model.} We employ the reward-model-deberta-v3-large-v2 \citep{openassistant_reward_model_2023} to score conversations. This model is trained on four diverse human-feedback datasets \citep{nakano2021webgpt,stienon2020learning,alex_havrilla_2023,bai2022training}, enabling it to perform evaluation on models' responses. We concatenate each entire multi-turn dialogue into a single input sequence and prompt the model to assign a reward score which reflects the overall quality. The score of a dataset is defined as the average reward score.\par
Table \ref{tab:Datasets Evaluation} presents the evaluation results. To derive an overall quality score for each dataset, we scale the \textit{Information Density} score by a factor of 100 and sum it with the other four evaluation metrics. The overall quality of the datasets is categorized as high (ChatAlpaca, M2Lingual, UltraChat), normal (WildChat, shareGPT), and low (OpenAssistant). 

\subsection{Experimental Settings}
\textbf{Parameter.} The experiments are conducted with instruct-style models from multiple families, including \textbf{LLaMA-3.2-3B-Instruct}, \textbf{LLaMA-3.1-8B-Instruct} \cite{llama3.2-3b-instruct}, \textbf{Qwen2.5-3B-Instruct}, and \textbf{Qwen2.5-7B-Instruct} \citep{qwen2024qwen25}. The models are fine-tuned for 3 epochs on datasets of varying quality. Each device processed a batch size of 4, with a gradient accumulation step of 4, resulting in an effective batch size of 64. The Adam optimizer was employed, with the hyperparameter $\beta_2$ set to 0.95. A cosine decay learning rate schedule was applied, starting at an initial learning rate of $1 \times 10^{-5}$ and incorporating a warm-up ratio of 0.01. All training and evaluation procedures were performed in FP16 precision on four NVIDIA GPUs. To reduce memory consumption, gradient checkpointing and Low-Rank Adaptation (LoRA) were enabled during training. Model performance was periodically assessed using a held-out validation set of 400 examples.

To enhance the robustness of the training process, a warm-up strategy was implemented during the initial phase of training. This involved using 640 high-quality dialogue samples to initialize baseline mean and variance parameters. As training progressed, the filtering weight for anomalous data was gradually increased to ensure smooth and stable model optimization.

\textbf{Evaluation.} We conducted evaluations across three settings: \textbf{In-Domain-Test}, \textbf{MT-Bench} \citep{zheng2023judging}, and \textbf{MT-Bench-Ext} \citep{kwan-etal-2024-mt}, to assess both in-domain performance and generalization. The In-Domain-Test serves as a setting-specific evaluation, where models are tested on held-out samples from the same distribution as the training data. It includes six multi-turn dialogue datasets (ShareGPT, WildChat, OpenAssistant, ChatAlpaca, M2Lingual, and UltraChat), each with 100 randomly sampled conversations to cover diverse domains and supervision styles. All evaluations followed the GPT-4–based “LLM-as-a-Judger” protocol \citep{zheng2023judging}, which was used both to compute Win Rate \citep{alpaca_eval, dubois2024length, dubois2023alpacafarm} via pairwise comparisons and to assign fine-grained scores across four human-aligned criteria: \textbf{Faithfulness} (Faith.), \textbf{Appropriateness} (Appr.), \textbf{Naturalness} (Nat.), and \textbf{Completeness} (Compl.).

\textbf{Mix Dataset.} To validate the effectiveness of our approach, we selected ChatAlpaca, ShareGPT, and OpenAssistant as representatives of high-, normal-, and low-quality datasets, respectively. From each dataset, 20K samples are extracted and mixed in different combinations: high and normal quality, high and low quality, and high, normal, and low quality. These experiments are designed to assess the performance of our method in handling datasets with varying distributions during training.

\begin{table*}[t!]
\renewcommand{\arraystretch}{1.1}  % 将行间距增大到原来的1.2倍
\centering
\scriptsize
\setlength{\tabcolsep}{0.5mm}{%
\begin{tabular}{>{\centering\arraybackslash}p{0.06\linewidth}>{\raggedright\arraybackslash}p{0.09\linewidth}>{\centering\arraybackslash}p{0.03\linewidth}>{\centering\arraybackslash}p{0.115\linewidth}>{\centering\arraybackslash}p{0.115\linewidth}>{\centering\arraybackslash}p{0.03\linewidth}>{\centering\arraybackslash}p{0.1\linewidth}>{\centering\arraybackslash}p{0.115\linewidth}>{\centering\arraybackslash}p{0.03\linewidth}>{\centering\arraybackslash}p{0.11\linewidth}>{\centering\arraybackslash}p{0.115\linewidth}}
\toprule
\multirow{2}{*}{\textbf{Level}}& \multirow{2}{*}{\textbf{Dataset}}& \multicolumn{3}{c}{\textbf{In-Domain-Test}} 
& \multicolumn{3}{c}{\textbf{MT-Bench}} 
& \multicolumn{3}{c}{\textbf{MT-Bench-Ext}} \\
\cmidrule(lr){3-5} \cmidrule(lr){6-8} \cmidrule(lr){9-11}
& & \textbf{BM}& \textbf{VT}& \textbf{ReSURE}
& \textbf{BM}& \textbf{VT}& \textbf{ReSURE}
& \textbf{BM}& \textbf{VT}& \textbf{ReSURE} \\
\midrule
\multirow{3}{*}{\textbf{H}}& M2Lin.& \textbf{7.10} & 7.09 (\textcolor{DarkRed}{-0.14\%}) & 7.06 (\textcolor{DarkRed}{-0.56\%})& 7.13 & \textbf{7.21} (\textcolor{DarkGreen}{+1.12\%}) & 7.16 (\textcolor{DarkGreen}{+0.42\%})
& 6.64 & 6.65 (\textcolor{DarkGreen}{+0.15\%}) & \textbf{6.71} (\textcolor{DarkGreen}{+1.05\%}) \\
& ChatAlpaca& 8.20 & 7.99 (\textcolor{DarkRed}{-2.56\%}) & \textbf{8.26} (\textcolor{DarkGreen}{+0.73\%})
& 7.13 & 6.97 (\textcolor{DarkRed}{-2.24\%}) & \textbf{7.29} (\textcolor{DarkGreen}{+2.24\%})
& 6.64 & 5.99 (\textcolor{DarkRed}{-9.79\%}) & \textbf{6.76} (\textcolor{DarkGreen}{+1.81\%}) \\
& UltraChat& 7.90 & 7.56 (\textcolor{DarkRed}{-4.30\%}) & \textbf{8.01} (\textcolor{DarkGreen}{+1.39\%})
& 7.13 & 6.68 (\textcolor{DarkRed}{-6.31\%}) & \textbf{7.32} (\textcolor{DarkGreen}{+2.66\%})
& 6.64 & 6.22 (\textcolor{DarkRed}{-6.33\%}) & \textbf{6.76} (\textcolor{DarkGreen}{+1.81\%}) \\
\hdashline
\multirow{2}{*}{\textbf{N}}& ShareGPT& 6.55 & 6.09 (\textcolor{DarkRed}{-7.02\%}) & \textbf{6.95} (\textcolor{DarkGreen}{+6.11\%})
& 7.13 & 6.08 (\textcolor{DarkRed}{-14.73\%}) & \textbf{7.83} (\textcolor{DarkGreen}{+9.82\%})
& 6.64 & 5.80 (\textcolor{DarkRed}{-12.65\%}) & \textbf{6.83} (\textcolor{DarkGreen}{+2.86\%}) \\
& WildChat& 6.80 & 6.47 (\textcolor{DarkRed}{-4.85\%}) & \textbf{6.86} (\textcolor{DarkGreen}{+0.88\%})
& 7.13 & 7.14 (\textcolor{DarkGreen}{+0.14\%}) & \textbf{7.21} (\textcolor{DarkGreen}{+1.12\%})
& 6.64 & 6.74 (\textcolor{DarkGreen}{+1.51\%}) & \textbf{6.72} (\textcolor{DarkGreen}{+1.20\%}) \\
\hdashline
\multirow{1}{*}{\textbf{L}}& OpenAss.& 7.64 & 7.20 (\textcolor{DarkRed}{-5.76\%}) & \textbf{7.67} (\textcolor{DarkGreen}{+0.39\%})
& 7.13 & 6.20 (\textcolor{DarkRed}{-13.07\%}) & \textbf{7.26} (\textcolor{DarkGreen}{+1.83\%})
& 6.64 & 5.48 (\textcolor{DarkRed}{-17.47\%}) & \textbf{6.83} (\textcolor{DarkGreen}{+2.86\%}) \\
\bottomrule
\end{tabular}
}
\caption{Comparison of our method, non-trained Base Model (BM), and Vicuna-Tuning on LLaMA-3.2-3B-Instruct: multi-turn dialogue performance (GPT-4 scores) across high-, normal-, and low-quality datasets. Each cell shows absolute scores plus relative improvement/decline (\%) vs.\ BM in parentheses. H, N, L = High, Normal, Low, M2Lin. = M2Lingual (en), OpenAss. = OpenAssistant.}
\label{tab:Main Experiment}
\end{table*}

\begin{figure*}[t!]
  \centering
  \includegraphics[width=\linewidth]{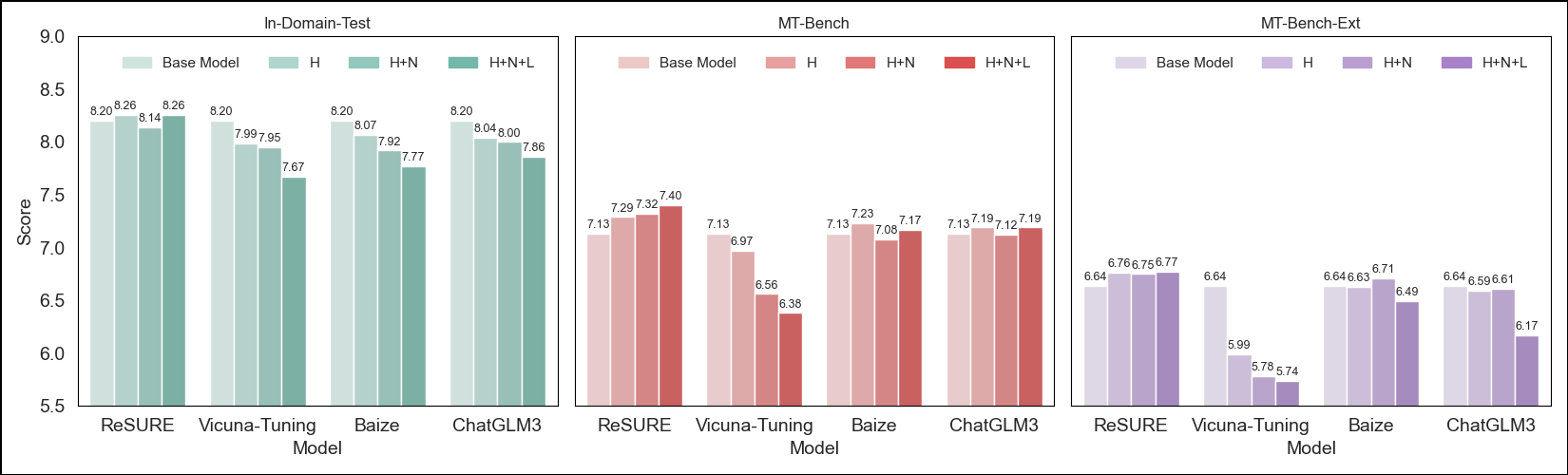}
  \caption{Performance scaling with Hierarchical Data Integration (H, H+N, H+N+L): (a) In-Domain-Test Performance (b) MT-Bench Performance (c) MT-Bench-Ext Performance}
  \label{f-mix-evaluation}
\end{figure*}

\subsection{Baselines}
We evaluate our method against four typical methods in multi-turn dialogue study: \\
(1) \textbf{Base Model (BM)}: the original instruction-tuned model without task-specific fine-tuning for multi-turn dialogue. \\
(2) \textbf{Vicuna-Tuning (VT)}: a widely adopted dialogue adaptation framework built upon LLaMA, distinguished by its LoRA fine-tuning strategy on multi-turn conversational data \cite{vicuna2023}. \\
(3) \textbf{Baize}: a parameter-efficient approach that exclusively updates linear layers through self-chat generation \cite{vicuna2023}. \\
(4) \textbf{ChatGLM3}: implements multi-turn dialogue fine-tuning by updating only the loss of roles other than \textit{user} and \textit{system} \cite{glm2024chatglm}.\\
All methods share identical LoRA configurations (rank=128, alpha=16, dropout=0.3) and data partitions: 20,000 training samples with 400 validation and 100 test instances. Experiments are conducted with fixed random seeds (seed=42) and multi-turn dialogue performance quantified by the MT-Bench \cite{zheng2023judging}.

\subsection{Main Results}

\subsubsection{{Does ReSURE address negative optimization in multi-turn dialogues?}}
To evaluate the ability of ReSURE to mitigate negative optimization, which refers to performance degradation as the volume of supervision increases, we conduct a comparative analysis with Vicuna-Tuning across six instruction-tuned multi-turn dialogue datasets: M2Lingual, ChatAlpaca, UltraChat, ShareGPT, Wildchat, and OpenAssistant. As shown in Table~\ref{tab:Main Experiment}, ReSURE consistently outperforms the base model by 6.11\%, 9.82\%, and 2.86\% on the in-domain benchmark, MT-Bench, and MT-Bench-Ext, respectively. In contrast, Vicuna-Tuning exhibits clear signs of negative optimization, particularly on ShareGPT, where additional training data reduces performance—likely due to stylistic inconsistencies or supervision conflicts. Although ReSURE achieves slightly lower gains on M2Lingual, this may be attributed to the limited dataset size and increased risk of overfitting. Overall, these results demonstrate that ReSURE scales effectively with increasing data while maintaining robustness to supervision noise.

Human evaluation on MT-Bench-Ext (Table~\ref{tab:human_eval_summary}) further supports our findings. ReSURE outperforms both the base model and Vicuna-Tuning across all evaluation dimensions, with notable improvements in Faithfulness and Completeness. These gains are especially evident in multi-turn settings, where maintaining factual consistency and contextual coherence is essential. The results indicate that ReSURE more effectively preserves semantic alignment across turns, resulting in more coherent and informative dialogues. Additional evaluation details are provided in Appendix~\ref{app:human_evaluation}.

These results indicate that the dynamic suppression of unreliable supervision contributes to more stable training dynamics and semantically aligned responses. This observation is consistent with the findings from automatic benchmarks, and further supports the robustness of ReSURE under imperfect supervision conditions in instruction-tuned dialogue settings.

\subsubsection{Does ReSURE Suppress Unreliable Supervision for Robust Fine-Tuning?}

% \tanfeicomment{}{this study attemps to demonstrate our method can work as junk/low-quality data filtering/skipping on the fly, thus experiments results are supposed to highlight it's positive with more data while others do the other way around. NOT JUST SAY "WE HAVE THE PERFORMANCE IMPROVEMENT AGAINST OTHER METHODS"!!}

To evaluate ReSURE’s robustness under noisy supervision, we construct mixed datasets of increasing complexity and compare it with Vicuna-Tuning, Baize, and ChatGLM3. This setup simulates realistic fine-tuning scenarios involving low-quality or off-distribution samples. As shown in Figure~\ref{f-mix-evaluation}, ReSURE maintains or improves performance across all three evaluation settings as dataset size and noise increase. It achieves stable in-domain scores around 8.2, with steady gains on MT-Bench (7.13 to 7.4) and MT-Bench-Ext (6.64 to 6.77), indicating effective use of additional supervision without overfitting to noise. In contrast, Vicuna-Tuning exhibits consistent degradation—particularly on MT-Bench-Ext (6.64→5.74)—while Baize and ChatGLM3 show marginal or unstable changes. These trends are confirmed by Spearman correlation analysis (Appendix Table~\ref{tab:setting_corr_spearman}), where ReSURE yields positive correlations across all benchmarks, unlike the negative or inconsistent values observed for baselines.

\textbf{ReSURE excels on partially noisy datasets, maintaining positive optimization.} As shown in Figure~X, when noise increases from high-quality (H) to mixed-quality (H+N+L), conventional methods like Vicuna-Tuning and Baize exhibit noticeable performance drops—e.g., Vicuna-Tuning drops by 0.75 on MT-Bench and 0.90 on MT-Bench-Ext. In contrast, ReSURE shows strong robustness, with minimal variance and even slight improvements in noisy conditions. On the In-Domain-Test, ReSURE achieves a peak score of 8.26, maintaining a high level of performance across all mixtures. In multi-turn settings, it consistently outperforms baselines across all noise levels, particularly under challenging H+N+L configurations. This resilience enables ReSURE to leverage larger and more diverse training data effectively, without requiring explicit pre-filtering.

\begin{table}[t]
\centering
\small
\renewcommand{\arraystretch}{1}
\begin{tabular}{lccccc}
\toprule
\textbf{Model} & \textbf{Faith.} & \textbf{Appr.} & \textbf{Nat.} & \textbf{Compl.} & \textbf{Over.} \\
\midrule
BM    & 3.40 & 3.06 & 3.16 & 3.50 & 3.04 \\
VT  & 3.36 & 2.98 & 3.22 & 3.42 & 2.98 \\
ReSURE   & \textbf{3.74} & \textbf{3.68} & \textbf{3.74} & \textbf{3.86} & \textbf{3.66} \\
\bottomrule
\end{tabular}
\caption{
Human evaluation on MT-Bench-Ext.
}
\label{tab:human_eval_summary}
\end{table}

\section{Ablation Study}

% Our ablation study examines the impact of mixing different types of datasets on model performance. The key finding is that our method maintains stability across all evaluation metrics, while Vicuna\_tuning suffers a slight decline when additional data sources are introduced.
% As shown in Figure.\ref{tab:comparison}, on the ChatAlpaca dataset, Ours achieves an in-domain test score of 8.24, an MT-bench score of 7.2, and an MT-Bench-Ext score of 6.76, while Vicuna\_tuning scores 7.99, 6.97, and 6.00, respectively. After incorporating GSM8K, our method retains stable performance across all metrics (8.20 in in-domain test, 7.37 in MT-bench, and 6.8 in MT-Bench-Ext), whereas Vicuna\_tuning exhibits a slight decrease (7.98, 6.82, and 5.92).

\newcolumntype{L}[1]{>{\raggedright\arraybackslash}m{#1}} 
\newcolumntype{P}[1]{>{\centering\arraybackslash}m{#1}}  % 定义居中列

\begin{table}[t]
\centering
\small
\renewcommand{\arraystretch}{1.1}  % 增大行间距
\begin{tabular}{L{1.43cm} P{1.55cm} P{1.43cm} P{1.5cm}}
\toprule
\textbf{Exp Setting} & \textbf{In-Domain-Test} & \textbf{MT-Bench} & \textbf{MT-Bench-Ext} \\
\midrule
Base & 8.20 & 7.13 & 6.64 \\
\hdashline
VT & 7.99 & 6.97 & 5.99 \\
VT + Prefiltering & 7.99 (\textcolor{black}{0.00\%}) & 7.15 (\textcolor{DarkGreen}{+2.58\%}) & 6.68 (\textcolor{DarkGreen}{+11.52\%}) \\
\hdashline
ReSURE & 8.26 & 7.29 & 6.76 \\
ReSURE + Prefiltering & \textbf{8.28} (\textcolor{DarkGreen}{+0.24\%}) & \textbf{7.58} (\textcolor{DarkGreen}{+3.98\%}) & \textbf{7.35} (\textcolor{DarkGreen}{+8.73\%}) \\
\bottomrule
\end{tabular}
\caption{Performance comparison between the prefiltering method (DeBERTa) and ReSURE.}
\label{tab:pre_filtering_ReSURE}
\end{table}

\subsection{Can ReSURE Handle Task Mixture?} 
\begin{figure}[t!]
  \centering
  \includegraphics[width=\linewidth]{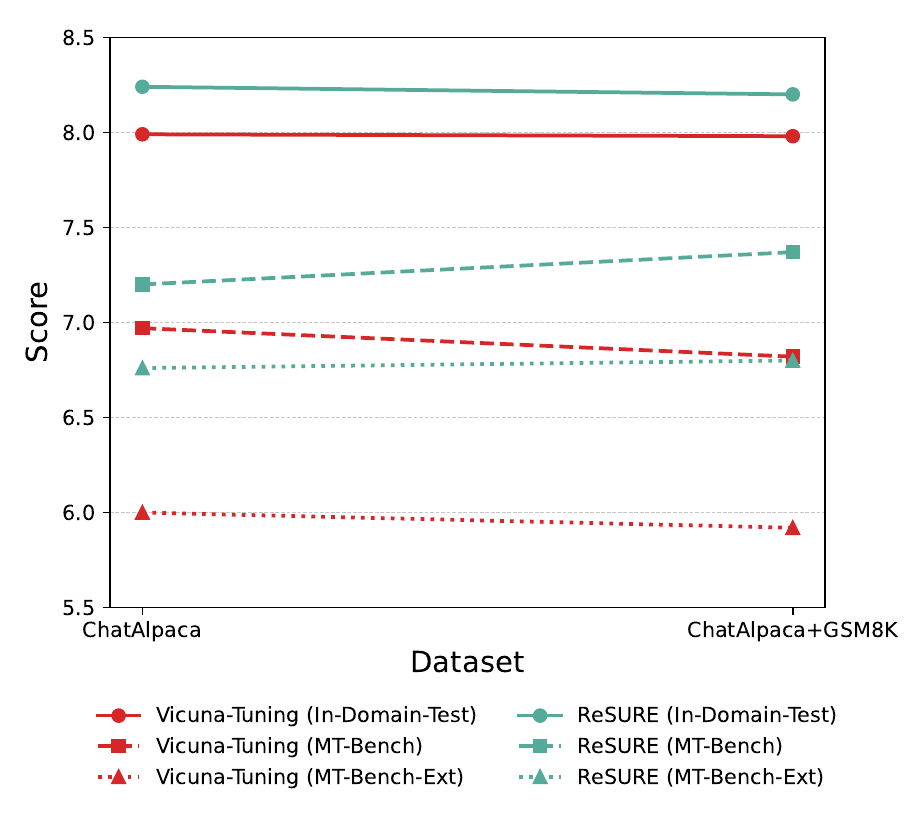}
  \caption{Performance comparison between Vicuna-Tuning and ReSURE on ChatAlpaca and ChatAlpaca+GSM8K across three evaluation benchmarks: In-Domain-Test, MT-Bench, and MT-Bench-Ext.}
  \label{f-noise_compare}
\end{figure}

To further examine the stability of ReSURE under heterogeneous training conditions, we incorporate GSM8K, a mathematical question answering dataset, into the multi-turn ChatAlpaca corpus. This setting introduces task-level noise due to divergent supervision styles. As shown in Figure~\ref{f-noise_compare}, ReSURE maintains in-domain performance and achieves positive generalization on MT-Bench and MT-Bench-Ext, even when trained on mixed-task data. In contrast, Vicuna-Tuning shows performance degradation on both in-domain and general benchmarks, likely due to overfitting to arithmetic patterns in GSM8K, which weakens its multi-turn dialogue capability and harms contextual alignment. These results indicate that ReSURE is more robust to task drift and better preserves dialogue-relevant optimization signals by dynamically suppressing incompatible supervision. All experiments are conducted using the \texttt{LLaMA3.2-3B-Instruct} model. Notably, ReSURE also improves GSM8K accuracy from 77.7\% to 78.3\%, confirming its robustness across tasks without sacrificing task-specific performance.

\begin{table}[t]
\centering
\small
\renewcommand{\arraystretch}{1}
\begin{tabular}{L{1.2cm} P{1.6cm} P{1.6cm} P{1.6cm}}
\toprule
\textbf{Model} & \textbf{In-Domain-Test} & \textbf{MT-Bench} & \textbf{MT-Bench-Ext} \\
\midrule
VT & -1.000 & -1.000 & -1.000 \\
Baize & -1.000 & 0.000 & -0.400 \\
ChatGLM3 & -1.000 & 0.211 & -0.800 \\
\hdashline
ReSURE & 0.211 & 1.000 & 0.800 \\
\bottomrule
\end{tabular}
\caption{Spearman correlation between dataset complexity and performance across benchmarks.}
\label{tab:setting_corr_spearman}
\end{table}

\subsection{{Does ReSURE perform better than pre-filtering methods?}}
We compare ReSURE against traditional offline reward-based pre-filtering \citep{du2023mods}, using reward-model-deberta-v3-large-v2 \citep{openassistant_reward_model_2023} to retain the top 75\% of samples from ChatAlpaca, ShareGPT, and OpenAssistant. As shown in Table~\ref{tab:pre_filtering_ReSURE}, ReSURE alone outperforms static filtering, and the best performance is achieved by combining both. Notably, this hybrid setup yields the largest improvement on MT-Bench-Ext, highlighting its advantage in complex multi-turn scenarios. These findings indicate that ReSURE’s dynamic reweighting complements static quality filtering, offering an effective synergy for robust dialogue fine-tuning.

\begin{figure}[t!]
  \centering
  \includegraphics[width=\linewidth]{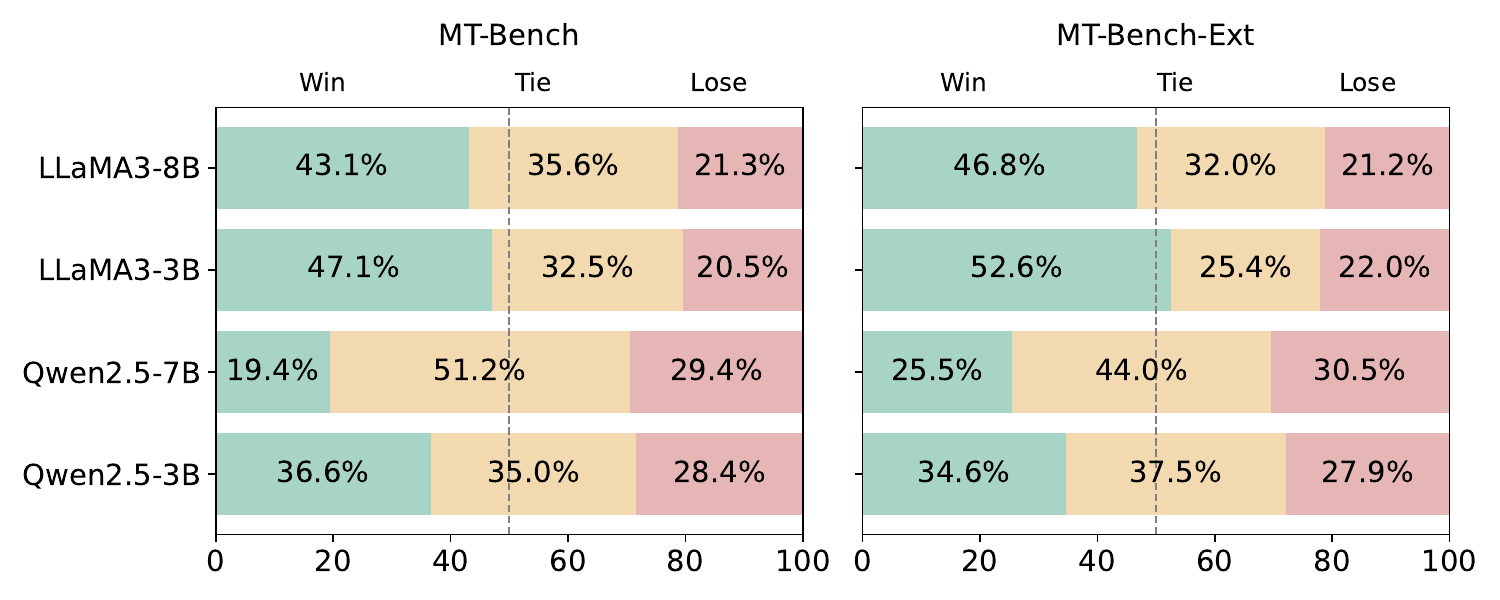}
  \caption{Win Rates of ReSURE vs. VT on MT-Bench and MT-Bench-Ext.}
  \label{f-ablation_res}
\end{figure}

\newcolumntype{L}[1]{>{\raggedright\arraybackslash}m{#1}} 
\newcolumntype{P}[1]{>{\centering\arraybackslash}m{#1}}  % 定义居中列

\begin{table}[t!]
\renewcommand{\arraystretch}{1.1}
\small
\centering
\setlength{\tabcolsep}{2.3mm}{
\begin{tabular}{lccc}
\hline
\textbf{Metric} & \textbf{ReSURE} & \makecell{\textbf{w/o Welford}} & \textbf{$\Delta$ (\%)} \\
\hline
In-Domain-Test     & 8.26 & 8.20 & \textbf{-0.73\%} \\
MT-Bench      & 7.29 & 7.19 & \textbf{-1.37\%} \\
MT-Bench-Ext  & 6.76 & 6.70 & \textbf{-0.89\%} \\
\hline
\end{tabular}
}
\caption{Performance drop (GPT-4 scores) when removing Welford statistics from ReSURE across three evaluation benchmarks.}
\label{tab:comparison}
\end{table}

\subsection{How Does ReSURE Perform Across Diverse Model Families and Sizes?}

To assess ReSURE’s generalizability, we apply ReSURE to four instruction-tuned models from the Qwen and LLaMA families. We evaluate using \textit{Win Rate}—the proportion of multi-turn responses preferred over base outputs, as judged by GPT-4. As shown in Figure~\ref{f-ablation_res}, ReSURE consistently improves multi-turn quality across settings. Improvements are more stable for LLaMA models, whereas Qwen models exhibit greater variance between MT-Bench and MT-Bench-Ext. This variance likely reflects the extensive pretraining of Qwen on high-quality multi-task instruction data, which enhances zero-shot and few-shot ability but also increases sensitivity to benchmark artifacts \citep{wu2025reasoning}. Despite this variability, ReSURE demonstrates robust improvements across both model families, highlighting its effectiveness and broad applicability.

% This variability likely stems from Qwen’s extensive pretraining on high-quality, web-scale multi-task instruction data .

% while Qwen models show greater variance between MT-Bench and MT-Bench-Ext, suggesting model-specific sensitivity to noisy supervision. These results highlight ReSURE’s applicability and robustness across architectures and scales.

\textbf{ReSURE enhances response performance by effectively skipping low-quality data.} To better understand the impact of its adaptive weighting mechanism, we conduct an ablation study by removing the Welford-based loss modulation, while keeping all other training settings and loss components unchanged. This ablated variant disables the skip mechanism and treats all supervision equally, regardless of quality. Table~\ref{tab:comparison} demonstrates that removing Welford statistics leads to performance drops of 0.73\%, 1.39\%, and 0.90\% on In-Domain, MT-Bench, and MT-Bench-Ext, respectively. These results confirm that selectively down-weighting unreliable supervision improves robustness and training stability in multi-turn dialogue tuning, and highlight the importance of adaptive loss modulation in mitigating the impact of noisy or inconsistent annotations.

\begin{figure}[t!]
  \centering
  \includegraphics[width=\linewidth]{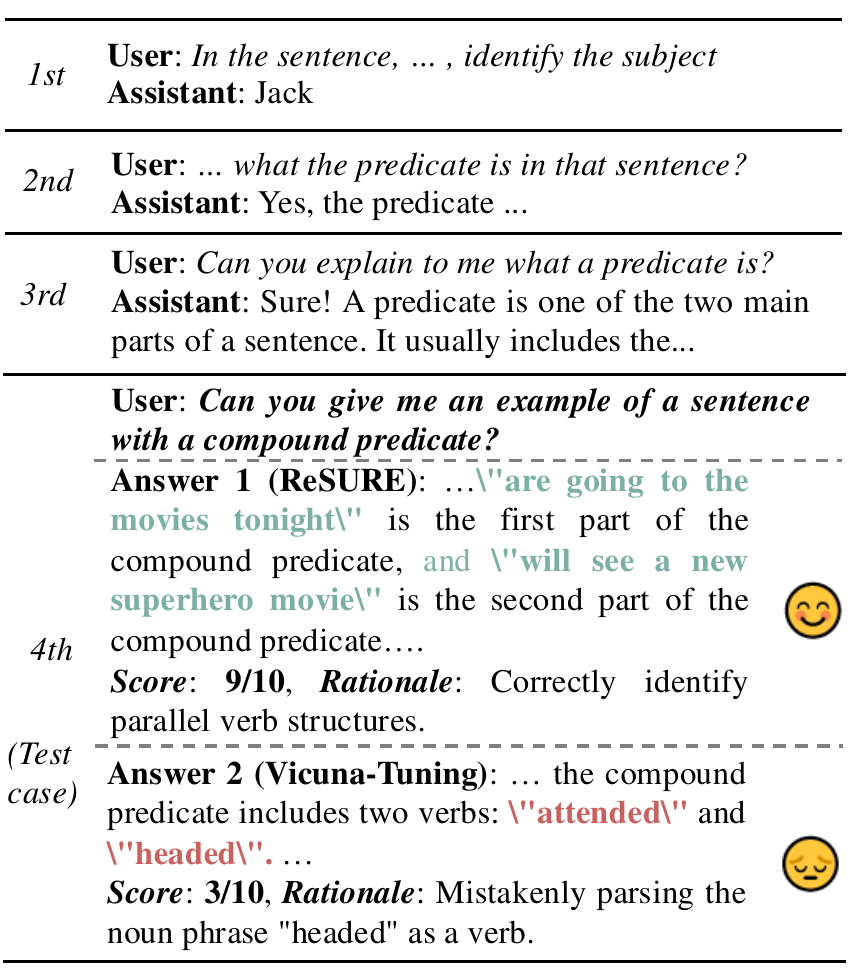}
  \caption{Case study.}
  \label{f-case_study-abb}
\end{figure}

\section{Case Study}

As illustrated in Figure~\ref{f-case_study-abb}, this multi-turn dialogue example demonstrates the superior contextual understanding of ReSURE compared to Vicuna-Tuning. When processing a compound predicate query, ReSURE correctly identifies the parallel verb structure, accurately parsing both predicate components ("are going" and "will see") with appropriate syntactic boundaries. In contrast, Vicuna-Tuning misinterprets the noun phrase "headed" as a verb predicate, despite the prior context clearly indicating "head" as a positional noun. This error highlights the model's limited ability to maintain dialogue state awareness and resolve referential dependencies across turns. Additional examples are provided in Appendix~\ref{app:case_study}.

\section{Conclusion}

We propose \textbf{ReSURE}, a turn-aware fine-tuning framework that dynamically down-weights unreliable supervision via per-turn loss statistics. Without explicit data filtering, ReSURE improves response quality and training stability across MT-Bench, MT-Bench-Ext, and in-domain settings. It demonstrates consistent gains under supervision noise, with ablations confirming the effectiveness of turn-aware modulation. ReSURE offers a scalable solution for instruction tuning on large, mixed-quality datasets.

\section*{Limitation}
This study has several limitations. First, while we adopt one type of online statistical approach, alternative techniques for modeling supervision reliability remain unexplored. Second, our dataset quality evaluation is intended as a reference rather than a definitive measure, as different domains may require tailored metrics. Third, the method is evaluated only in multi-turn dialogue scenarios, with broader applications limited by computational cost. In addition, our results on Qwen2.5-7B-Instruct are less promising compared to other models, potentially due to architectural differences or instruction tuning strategies not well aligned with our loss calibration mechanism. Despite these limitations, we hope our findings offer useful insights for future research on domain-specific fine-tuning.

\section*{Ethics Statement}
This research focuses on improving the robustness of fine-tuning multi-turn dialogue systems using publicly available datasets. All datasets used in this work are released under permissive licenses and do not contain personally identifiable information. No human subjects were involved in data collection. While our method aims to reduce the impact of unreliable supervision, it implicitly filters training signals, which may lead to unintended bias or underrepresentation of minority styles. Model evaluations are conducted by three trained research assistants, each paid $\$$20/hour, above the local average.
\section*{Acknowledgements}
This work is partially supported by Hong Kong RGC GRF No. 14206324 and CUHK Knowledge Transfer Project Fund No. KPF23GWP20. This research was also supported
by SenseTime.
% \newpage
% Entries for the entire Anthology, followed by custom entries
\bibliography{acl_latex}

\begin{thebibliography}{74}
\expandafter\ifx\csname natexlab\endcsname\relax\def\natexlab#1{#1}\fi

\bibitem[{Bai et~al.(2024)Bai, Liu, Bu, He, Liu, Zhou, Lin, Su, Ge, Zheng, and Ouyang}]{bai-etal-2024-mt}
Ge~Bai, Jie Liu, Xingyuan Bu, Yancheng He, Jiaheng Liu, Zhanhui Zhou, Zhuoran Lin, Wenbo Su, Tiezheng Ge, Bo~Zheng, and Wanli Ouyang. 2024.
\newblock \href {https://doi.org/10.18653/v1/2024.acl-long.401} {{MT}-bench-101: A fine-grained benchmark for evaluating large language models in multi-turn dialogues}.
\newblock In \emph{Proceedings of the 62nd Annual Meeting of the Association for Computational Linguistics (Volume 1: Long Papers)}, pages 7421--7454, Bangkok, Thailand. Association for Computational Linguistics.

\bibitem[{Bai et~al.(2022)Bai, Jones, Ndousse, Askell, Chen, DasSarma, Drain, Fort, Ganguli, Henighan et~al.}]{bai2022training}
Yuntao Bai, Andy Jones, Kamal Ndousse, Amanda Askell, Anna Chen, Nova DasSarma, Dawn Drain, Stanislav Fort, Deep Ganguli, Tom Henighan, et~al. 2022.
\newblock Training a helpful and harmless assistant with reinforcement learning from human feedback.
\newblock \emph{arXiv preprint arXiv:2204.05862}.

\bibitem[{Bian et~al.(2023)Bian, Lin, Lu, Han, Sun, and He}]{ChatAlpaca}
Ning Bian, Hongyu Lin, Yaojie Lu, Xianpei Han, Le~Sun, and Ben He. 2023.
\newblock Chatalpaca: A multi-turn dialogue corpus based on alpaca instructions.
\newblock \url{https://github.com/cascip/ChatAlpaca}.

\bibitem[{Cao et~al.(2023)Cao, Kang, Wang, and Sun}]{cao2023instruction}
Yihan Cao, Yanbin Kang, Chi Wang, and Lichao Sun. 2023.
\newblock Instruction mining: Instruction data selection for tuning large language models.
\newblock \emph{arXiv preprint arXiv:2307.06290}.

\bibitem[{Chen et~al.(2023)Chen, Li, Yan, Wang, Gunaratna, Yadav, Tang, Srinivasan, Zhou, Huang et~al.}]{chen2023alpagasus}
Lichang Chen, Shiyang Li, Jun Yan, Hai Wang, Kalpa Gunaratna, Vikas Yadav, Zheng Tang, Vijay Srinivasan, Tianyi Zhou, Heng Huang, et~al. 2023.
\newblock Alpagasus: Training a better alpaca with fewer data.
\newblock \emph{arXiv preprint arXiv:2307.08701}.

\bibitem[{Chiang et~al.(2023)Chiang, Li, Lin, Sheng, Wu, Zhang, Zheng, Zhuang, Zhuang, Gonzalez, Stoica, and Xing}]{vicuna2023}
Wei-Lin Chiang, Zhuohan Li, Zi~Lin, Ying Sheng, Zhanghao Wu, Hao Zhang, Lianmin Zheng, Siyuan Zhuang, Yonghao Zhuang, Joseph~E. Gonzalez, Ion Stoica, and Eric~P. Xing. 2023.
\newblock Vicuna: An open-source chatbot impressing gpt-4 with 90\%* chatgpt quality.
\newblock \url{https://lmsys.org/blog/2023-03-30-vicuna/}.
\newblock Accessed: 2025-02-10.

\bibitem[{Cobbe et~al.(2021)Cobbe, Kosaraju, Bavarian, Chen, Jun, Kaiser, Plappert, Tworek, Hilton, Nakano et~al.}]{cobbe2021training}
Karl Cobbe, Vineet Kosaraju, Mohammad Bavarian, Mark Chen, Heewoo Jun, Lukasz Kaiser, Matthias Plappert, Jerry Tworek, Jacob Hilton, Reiichiro Nakano, et~al. 2021.
\newblock Training verifiers to solve math word problems, 2021.
\newblock \emph{URL https://arxiv. org/abs/2110.14168}, 9.

\bibitem[{Contributors(2023)}]{OpenAssistant}
OpenAssistant Contributors. 2023.
\newblock Openassistant conversations - democratizing large language model alignment.
\newblock \url{https://arxiv.org/abs/2304.07327}.
\newblock Accessed: 2023-04-17.

\bibitem[{Dethlefs et~al.(2016)Dethlefs, Hastie, Cuay\'{a}huitl, Yu, Rieser, and Lemon}]{10.1016/j.csl.2015.11.001}
Nina Dethlefs, Helen Hastie, Heriberto Cuay\'{a}huitl, Yanchao Yu, Verena Rieser, and Oliver Lemon. 2016.
\newblock \href {https://doi.org/10.1016/j.csl.2015.11.001} {Information density and overlap in spoken dialogue}.
\newblock \emph{Comput. Speech Lang.}, 37(C):82–97.

\bibitem[{Dettmers et~al.(2024)Dettmers, Pagnoni, Holtzman, and Zettlemoyer}]{dettmers2024qlora}
Tim Dettmers, Artidoro Pagnoni, Ari Holtzman, and Luke Zettlemoyer. 2024.
\newblock Qlora: Efficient finetuning of quantized llms.
\newblock \emph{Advances in Neural Information Processing Systems}, 36.

\bibitem[{Ding et~al.(2023)Ding, Chen, Xu, Qin, Zheng, Hu, Liu, Sun, and Zhou}]{ding2023enhancing}
Ning Ding, Yulin Chen, Bokai Xu, Yujia Qin, Zhi Zheng, Shengding Hu, Zhiyuan Liu, Maosong Sun, and Bowen Zhou. 2023.
\newblock Enhancing chat language models by scaling high-quality instructional conversations.
\newblock \emph{arXiv preprint arXiv:2305.14233}.

\bibitem[{Du et~al.(2023)Du, Zong, and Zhang}]{du2023mods}
Qianlong Du, Chengqing Zong, and Jiajun Zhang. 2023.
\newblock Mods: Model-oriented data selection for instruction tuning.
\newblock \emph{arXiv preprint arXiv:2311.15653}.

\bibitem[{Du et~al.(2025)Du, Wang, He, Liang, Wang, Li, Gui, Pan, Xu, and Wong}]{du2025bridging}
Yiming Du, Bingbing Wang, Yang He, Bin Liang, Baojun Wang, Zhongyang Li, Lin Gui, Jeff~Z Pan, Ruifeng Xu, and Kam-Fai Wong. 2025.
\newblock Bridging the long-term gap: A memory-active policy for multi-session task-oriented dialogue.
\newblock \emph{arXiv preprint arXiv:2505.20231}.

\bibitem[{Dubois et~al.(2024)Dubois, Galambosi, Liang, and Hashimoto}]{dubois2024length}
Yann Dubois, Bal{\'a}zs Galambosi, Percy Liang, and Tatsunori~B Hashimoto. 2024.
\newblock Length-controlled alpacaeval: A simple way to debias automatic evaluators.
\newblock \emph{arXiv preprint arXiv:2404.04475}.

\bibitem[{Dubois et~al.(2023)Dubois, Li, Taori, Zhang, Gulrajani, Ba, Guestrin, Liang, and Hashimoto}]{dubois2023alpacafarm}
Yann Dubois, Xuechen Li, Rohan Taori, Tianyi Zhang, Ishaan Gulrajani, Jimmy Ba, Carlos Guestrin, Percy Liang, and Tatsunori~B. Hashimoto. 2023.
\newblock \href {http://arxiv.org/abs/2305.14387} {Alpacafarm: A simulation framework for methods that learn from human feedback}.

\bibitem[{Efanov et~al.(2021)Efanov, Ivliev, and Shagraev}]{efanov2021welford}
Andrey~A Efanov, Sergey~A Ivliev, and Alexey~G Shagraev. 2021.
\newblock Welford’s algorithm for weighted statistics.
\newblock In \emph{2021 3rd International Youth Conference on Radio Electronics, Electrical and Power Engineering (REEPE)}, pages 1--5. IEEE.

\bibitem[{GLM et~al.(2024)GLM, Zeng, Xu, Wang, Zhang, Yin, Rojas, Feng, Zhao, Lai, Yu, Wang, Sun, Zhang, Cheng, Gui, Tang, Zhang, Li, Zhao, Wu, Zhong, Liu, Huang, Zhang, Zheng, Lu, Duan, Zhang, Cao, Yang, Tam, Zhao, Liu, Xia, Zhang, Gu, Lv, Liu, Liu, Yang, Song, Zhang, An, Xu, Niu, Yang, Li, Bai, Dong, Qi, Wang, Yang, Du, Hou, and Wang}]{glm2024chatglm}
Team GLM, Aohan Zeng, Bin Xu, Bowen Wang, Chenhui Zhang, Da~Yin, Diego Rojas, Guanyu Feng, Hanlin Zhao, Hanyu Lai, Hao Yu, Hongning Wang, Jiadai Sun, Jiajie Zhang, Jiale Cheng, Jiayi Gui, Jie Tang, Jing Zhang, Juanzi Li, Lei Zhao, Lindong Wu, Lucen Zhong, Mingdao Liu, Minlie Huang, Peng Zhang, Qinkai Zheng, Rui Lu, Shuaiqi Duan, Shudan Zhang, Shulin Cao, Shuxun Yang, Weng~Lam Tam, Wenyi Zhao, Xiao Liu, Xiao Xia, Xiaohan Zhang, Xiaotao Gu, Xin Lv, Xinghan Liu, Xinyi Liu, Xinyue Yang, Xixuan Song, Xunkai Zhang, Yifan An, Yifan Xu, Yilin Niu, Yuantao Yang, Yueyan Li, Yushi Bai, Yuxiao Dong, Zehan Qi, Zhaoyu Wang, Zhen Yang, Zhengxiao Du, Zhenyu Hou, and Zihan Wang. 2024.
\newblock \href {http://arxiv.org/abs/2406.12793} {Chatglm: A family of large language models from glm-130b to glm-4 all tools}.

\bibitem[{Grattafiori et~al.(2024)Grattafiori, Dubey, Jauhri, Pandey, Kadian, Al-Dahle, Letman, Mathur, Schelten, Vaughan et~al.}]{llama3.2-3b-instruct}
Aaron Grattafiori, Abhimanyu Dubey, Abhinav Jauhri, Abhinav Pandey, Abhishek Kadian, Ahmad Al-Dahle, Aiesha Letman, Akhil Mathur, Alan Schelten, Alex Vaughan, et~al. 2024.
\newblock The llama 3 herd of models.
\newblock \emph{arXiv preprint arXiv:2407.21783}.

\bibitem[{Gu et~al.(2024)Gu, Yang, Ding, Zhao, and Tan}]{gu-etal-2024-cmr}
Jiawei Gu, Zacc Yang, Chuanghao Ding, Rui Zhao, and Fei Tan. 2024.
\newblock \href {https://doi.org/10.18653/v1/2024.emnlp-main.903} {{CMR} scaling law: Predicting critical mixture ratios for continual pre-training of language models}.
\newblock In \emph{Proceedings of the 2024 Conference on Empirical Methods in Natural Language Processing}, pages 16143--16162, Miami, Florida, USA. Association for Computational Linguistics.

\bibitem[{Hase et~al.(2024)Hase, Bansal, Clark, and Wiegreffe}]{hase-etal-2024-unreasonable}
Peter Hase, Mohit Bansal, Peter Clark, and Sarah Wiegreffe. 2024.
\newblock \href {https://doi.org/10.18653/v1/2024.acl-long.378} {The unreasonable effectiveness of easy training data for hard tasks}.
\newblock In \emph{Proceedings of the 62nd Annual Meeting of the Association for Computational Linguistics (Volume 1: Long Papers)}, pages 7002--7024, Bangkok, Thailand. Association for Computational Linguistics.

\bibitem[{Havrilla(2023)}]{alex_havrilla_2023}
Alex Havrilla. 2023.
\newblock \href {https://doi.org/10.57967/hf/1428} {synthetic-instruct-gptj-pairwise (revision cc92d8d)}.

\bibitem[{He et~al.(2021)He, Gao, and Chen}]{he2021debertav3}
Pengcheng He, Jianfeng Gao, and Weizhu Chen. 2021.
\newblock Debertav3: Improving deberta using electra-style pre-training with gradient-disentangled embedding sharing.
\newblock \emph{arXiv preprint arXiv:2111.09543}.

\bibitem[{He et~al.(2020)He, Liu, Gao, and Chen}]{he2020deberta}
Pengcheng He, Xiaodong Liu, Jianfeng Gao, and Weizhu Chen. 2020.
\newblock Deberta: Decoding-enhanced bert with disentangled attention.
\newblock \emph{arXiv preprint arXiv:2006.03654}.

\bibitem[{Hu et~al.(2021)Hu, Shen, Wallis, Allen-Zhu, Li, Wang, Wang, and Chen}]{hu2021lora}
Edward~J Hu, Yelong Shen, Phillip Wallis, Zeyuan Allen-Zhu, Yuanzhi Li, Shean Wang, Lu~Wang, and Weizhu Chen. 2021.
\newblock Lora: Low-rank adaptation of large language models.
\newblock \emph{arXiv preprint arXiv:2106.09685}.

\bibitem[{Hu et~al.(2025)Hu, Robey, and Liu}]{hu2025steering}
Hanjiang Hu, Alexander Robey, and Changliu Liu. 2025.
\newblock Steering dialogue dynamics for robustness against multi-turn jailbreaking attacks.
\newblock \emph{arXiv preprint arXiv:2503.00187}.

\bibitem[{Hu et~al.(2023)Hu, Wang, Lan, Xu, Lim, Bing, Xu, Poria, and Lee}]{hu-etal-2023-llm}
Zhiqiang Hu, Lei Wang, Yihuai Lan, Wanyu Xu, Ee-Peng Lim, Lidong Bing, Xing Xu, Soujanya Poria, and Roy Lee. 2023.
\newblock \href {https://doi.org/10.18653/v1/2023.emnlp-main.319} {{LLM}-adapters: An adapter family for parameter-efficient fine-tuning of large language models}.
\newblock In \emph{Proceedings of the 2023 Conference on Empirical Methods in Natural Language Processing}, pages 5254--5276, Singapore. Association for Computational Linguistics.

\bibitem[{K{\"o}pf et~al.(2024)K{\"o}pf, Kilcher, von R{\"u}tte, Anagnostidis, Tam, Stevens, Barhoum, Nguyen, Stanley, Nagyfi et~al.}]{kopf2024openassistant}
Andreas K{\"o}pf, Yannic Kilcher, Dimitri von R{\"u}tte, Sotiris Anagnostidis, Zhi~Rui Tam, Keith Stevens, Abdullah Barhoum, Duc Nguyen, Oliver Stanley, Rich{\'a}rd Nagyfi, et~al. 2024.
\newblock Openassistant conversations-democratizing large language model alignment.
\newblock \emph{Advances in Neural Information Processing Systems}, 36.

\bibitem[{Kwan et~al.(2024{\natexlab{a}})Kwan, Zeng, Jiang, Wang, Li, Shang, Jiang, Liu, and Wong}]{kwan2024mt}
Wai-Chung Kwan, Xingshan Zeng, Yuxin Jiang, Yufei Wang, Liangyou Li, Lifeng Shang, Xin Jiang, Qun Liu, and Kam-Fai Wong. 2024{\natexlab{a}}.
\newblock Mt-eval: A multi-turn capabilities evaluation benchmark for large language models.
\newblock \emph{arXiv preprint arXiv:2401.16745}.

\bibitem[{Kwan et~al.(2024{\natexlab{b}})Kwan, Zeng, Jiang, Wang, Li, Shang, Jiang, Liu, and Wong}]{kwan-etal-2024-mt}
Wai-Chung Kwan, Xingshan Zeng, Yuxin Jiang, Yufei Wang, Liangyou Li, Lifeng Shang, Xin Jiang, Qun Liu, and Kam-Fai Wong. 2024{\natexlab{b}}.
\newblock \href {https://doi.org/10.18653/v1/2024.emnlp-main.1124} {{MT}-eval: A multi-turn capabilities evaluation benchmark for large language models}.
\newblock In \emph{Proceedings of the 2024 Conference on Empirical Methods in Natural Language Processing}, pages 20153--20177, Miami, Florida, USA. Association for Computational Linguistics.

\bibitem[{Lei et~al.(2025)Lei, Ji, and Liu}]{lei2025mining}
Yutian Lei, Luping Ji, and Pei Liu. 2025.
\newblock Mining in-distribution attributes in outliers for out-of-distribution detection.
\newblock In \emph{Proceedings of the AAAI Conference on Artificial Intelligence}, volume~39, pages 18181--18188.

\bibitem[{Li et~al.(2024{\natexlab{a}})Li, Zhang, Li, Chen, Chen, Cheng, Wang, Zhou, and Xiao}]{li-etal-2024-quantity}
Ming Li, Yong Zhang, Zhitao Li, Jiuhai Chen, Lichang Chen, Ning Cheng, Jianzong Wang, Tianyi Zhou, and Jing Xiao. 2024{\natexlab{a}}.
\newblock \href {https://doi.org/10.18653/v1/2024.naacl-long.421} {From quantity to quality: Boosting {LLM} performance with self-guided data selection for instruction tuning}.
\newblock In \emph{Proceedings of the 2024 Conference of the North American Chapter of the Association for Computational Linguistics: Human Language Technologies (Volume 1: Long Papers)}, pages 7602--7635, Mexico City, Mexico. Association for Computational Linguistics.

\bibitem[{Li et~al.(2023{\natexlab{a}})Li, Yu, Zhou, Schick, Levy, Zettlemoyer, Weston, and Lewis}]{li2023self}
Xian Li, Ping Yu, Chunting Zhou, Timo Schick, Omer Levy, Luke Zettlemoyer, Jason Weston, and Mike Lewis. 2023{\natexlab{a}}.
\newblock Self-alignment with instruction backtranslation.
\newblock \emph{arXiv preprint arXiv:2308.06259}.

\bibitem[{Li et~al.(2023{\natexlab{b}})Li, Zhang, Dubois, Taori, Gulrajani, Guestrin, Liang, and Hashimoto}]{alpaca_eval}
Xuechen Li, Tianyi Zhang, Yann Dubois, Rohan Taori, Ishaan Gulrajani, Carlos Guestrin, Percy Liang, and Tatsunori~B. Hashimoto. 2023{\natexlab{b}}.
\newblock Alpacaeval: An automatic evaluator of instruction-following models.
\newblock \url{https://github.com/tatsu-lab/alpaca_eval}.

\bibitem[{Li et~al.(2024{\natexlab{b}})Li, Hui, Xia, Yang, Yang, Zhang, Si, Chen, Liu, Liu, Huang, and Li}]{li-etal-2024-one}
Yunshui Li, Binyuan Hui, Xiaobo Xia, Jiaxi Yang, Min Yang, Lei Zhang, Shuzheng Si, Ling-Hao Chen, Junhao Liu, Tongliang Liu, Fei Huang, and Yongbin Li. 2024{\natexlab{b}}.
\newblock \href {https://doi.org/10.18653/v1/2024.acl-long.252} {One-shot learning as instruction data prospector for large language models}.
\newblock In \emph{Proceedings of the 62nd Annual Meeting of the Association for Computational Linguistics (Volume 1: Long Papers)}, pages 4586--4601, Bangkok, Thailand. Association for Computational Linguistics.

\bibitem[{Liu et~al.(2024{\natexlab{a}})Liu, Wang, Yin, Molchanov, Wang, Cheng, and Chen}]{liu2024dora}
Shih-Yang Liu, Chien-Yi Wang, Hongxu Yin, Pavlo Molchanov, Yu-Chiang~Frank Wang, Kwang-Ting Cheng, and Min-Hung Chen. 2024{\natexlab{a}}.
\newblock Dora: Weight-decomposed low-rank adaptation.
\newblock \emph{arXiv preprint arXiv:2402.09353}.

\bibitem[{Liu et~al.(2024{\natexlab{b}})Liu, Zeng, He, Jiang, and He}]{liu2024what}
Wei Liu, Weihao Zeng, Keqing He, Yong Jiang, and Junxian He. 2024{\natexlab{b}}.
\newblock \href {https://openreview.net/forum?id=BTKAeLqLMw} {What makes good data for alignment? a comprehensive study of automatic data selection in instruction tuning}.
\newblock In \emph{The Twelfth International Conference on Learning Representations}.

\bibitem[{Lu et~al.(2023{\natexlab{a}})Lu, Zhu, Han, Zhao, Mac~Namee, and Tan}]{lu-etal-2023-makes}
Jinghui Lu, Dongsheng Zhu, Weidong Han, Rui Zhao, Brian Mac~Namee, and Fei Tan. 2023{\natexlab{a}}.
\newblock \href {https://doi.org/10.18653/v1/2023.acl-long.128} {What makes pre-trained language models better zero-shot learners?}
\newblock In \emph{Proceedings of the 61st Annual Meeting of the Association for Computational Linguistics (Volume 1: Long Papers)}, pages 2288--2303, Toronto, Canada. Association for Computational Linguistics.

\bibitem[{Lu et~al.(2023{\natexlab{b}})Lu, An, Lin, Pergola, He, Yin, Sun, and Wu}]{lu2023memochat}
Junru Lu, Siyu An, Mingbao Lin, Gabriele Pergola, Yulan He, Di~Yin, Xing Sun, and Yunsheng Wu. 2023{\natexlab{b}}.
\newblock Memochat: Tuning llms to use memos for consistent long-range open-domain conversation.
\newblock \emph{arXiv preprint arXiv:2308.08239}.

\bibitem[{Lu et~al.(2023{\natexlab{c}})Lu, Yuan, Yuan, Lin, Lin, Tan, and Zhou}]{lu2023instag}
Keming Lu, Hongyi Yuan, Zheng Yuan, Runji Lin, Junyang Lin, Chuanqi Tan, and Chang Zhou. 2023{\natexlab{c}}.
\newblock \# instag: Instruction tagging for diversity and complexity analysis.
\newblock \emph{arXiv preprint arXiv:2308.07074}.

\bibitem[{Maheshwary et~al.(2024)Maheshwary, Yadav, Nguyen, Mahajan, and Madhusudhan}]{maheshwary2024m2lingual}
Rishabh Maheshwary, Vikas Yadav, Hoang Nguyen, Khyati Mahajan, and Sathwik~Tejaswi Madhusudhan. 2024.
\newblock M2lingual: Enhancing multilingual, multi-turn instruction alignment in large language models.
\newblock \emph{arXiv preprint arXiv:2406.16783}.

\bibitem[{Meng et~al.(2024)Meng, Wang, and Zhang}]{meng2024pissa}
Fanxu Meng, Zhaohui Wang, and Muhan Zhang. 2024.
\newblock Pissa: Principal singular values and singular vectors adaptation of large language models.
\newblock \emph{arXiv preprint arXiv:2404.02948}.

\bibitem[{Nakano et~al.(2021)Nakano, Hilton, Balaji, Wu, Ouyang, Kim, Hesse, Jain, Kosaraju, Saunders, Jiang, Cobbe, Eloundou, Krueger, Button, Knight, Chess, and Schulman}]{nakano2021webgpt}
Reiichiro Nakano, Jacob Hilton, Suchir Balaji, Jeff Wu, Long Ouyang, Christina Kim, Christopher Hesse, Shantanu Jain, Vineet Kosaraju, William Saunders, Xu~Jiang, Karl Cobbe, Tyna Eloundou, Gretchen Krueger, Kevin Button, Matthew Knight, Benjamin Chess, and John Schulman. 2021.
\newblock Webgpt: Browser-assisted question-answering with human feedback.
\newblock In \emph{arXiv}.

\bibitem[{OpenAI(2023)}]{openai2023gpt4}
OpenAI. 2023.
\newblock \href {https://arxiv.org/abs/2303.08774} {Gpt-4 technical report}.
\newblock \emph{arXiv preprint arXiv:2303.08774}.

\bibitem[{{OpenAssistant}(2023)}]{openassistant_reward_model_2023}
{OpenAssistant}. 2023.
\newblock Openassistant/reward-model-deberta-v3-large-v2.
\newblock \url{https://huggingface.co/OpenAssistant/reward-model-deberta-v3-large-v2}.
\newblock Reward model trained from human feedback to predict which generated answer is better judged by a human, given a question.

\bibitem[{Ou et~al.(2024)Ou, Wu, Liu, Zhang, Zhang, and Gai}]{ou-etal-2024-inductive}
Jiao Ou, Jiayu Wu, Che Liu, Fuzheng Zhang, Di~Zhang, and Kun Gai. 2024.
\newblock \href {https://doi.org/10.18653/v1/2024.emnlp-main.964} {Inductive-deductive strategy reuse for multi-turn instructional dialogues}.
\newblock In \emph{Proceedings of the 2024 Conference on Empirical Methods in Natural Language Processing}, pages 17402--17431, Miami, Florida, USA. Association for Computational Linguistics.

\bibitem[{Radziwill and Benton(2017)}]{radziwill2017evaluating}
Nicole~M Radziwill and Morgan~C Benton. 2017.
\newblock Evaluating quality of chatbots and intelligent conversational agents.
\newblock \emph{arXiv preprint arXiv:1704.04579}.

\bibitem[{Ren et~al.(2018)Ren, Zeng, Yang, and Urtasun}]{ren2018learning}
Mengye Ren, Wenyuan Zeng, Bin Yang, and Raquel Urtasun. 2018.
\newblock Learning to reweight examples for robust deep learning.
\newblock In \emph{International conference on machine learning}, pages 4334--4343. PMLR.

\bibitem[{RyokoAI(2023)}]{ShareGPT}
RyokoAI. 2023.
\newblock Sharegpt.
\newblock \url{https://huggingface.co/datasets/RyokoAI/ShareGPT52K}.

\bibitem[{Shani et~al.(2024)Shani, Rosenberg, Cassel, Lang, Calandriello, Zipori, Noga, Keller, Piot, Szpektor et~al.}]{shani2024multi}
Lior Shani, Aviv Rosenberg, Asaf Cassel, Oran Lang, Daniele Calandriello, Avital Zipori, Hila Noga, Orgad Keller, Bilal Piot, Idan Szpektor, et~al. 2024.
\newblock Multi-turn reinforcement learning from preference human feedback.
\newblock \emph{arXiv preprint arXiv:2405.14655}.

\bibitem[{Singh et~al.(2024)Singh, Vargus, Dsouza, Karlsson, Mahendiran, Ko, Shandilya, Patel, Mataciunas, OMahony et~al.}]{singh2024aya}
Shivalika Singh, Freddie Vargus, Daniel Dsouza, B{\"o}rje~F Karlsson, Abinaya Mahendiran, Wei-Yin Ko, Herumb Shandilya, Jay Patel, Deividas Mataciunas, Laura OMahony, et~al. 2024.
\newblock Aya dataset: An open-access collection for multilingual instruction tuning.
\newblock \emph{arXiv preprint arXiv:2402.06619}.

\bibitem[{Stiennon et~al.(2020)Stiennon, Ouyang, Wu, Ziegler, Lowe, Voss, Radford, Amodei, and Christiano}]{stienon2020learning}
Nisan Stiennon, Long Ouyang, Jeff Wu, Daniel~M. Ziegler, Ryan Lowe, Chelsea Voss, Alec Radford, Dario Amodei, and Paul Christiano. 2020.
\newblock Learning to summarize from human feedback.
\newblock In \emph{NeurIPS}.

\bibitem[{Sun et~al.(2024)Sun, Liu, Zhou, Huang, Song, Zhao, Zhang, Zhang, and Gai}]{sun-etal-2024-parrot}
Yuchong Sun, Che Liu, Kun Zhou, Jinwen Huang, Ruihua Song, Xin Zhao, Fuzheng Zhang, Di~Zhang, and Kun Gai. 2024.
\newblock \href {https://doi.org/10.18653/v1/2024.acl-long.525} {Parrot: Enhancing multi-turn instruction following for large language models}.
\newblock In \emph{Proceedings of the 62nd Annual Meeting of the Association for Computational Linguistics (Volume 1: Long Papers)}, pages 9729--9750, Bangkok, Thailand. Association for Computational Linguistics.

\bibitem[{Taori et~al.(2023)Taori, Gulrajani, Zhang, Dubois, Li, Guestrin, Liang, and Hashimoto}]{alpaca}
Rohan Taori, Ishaan Gulrajani, Tianyi Zhang, Yann Dubois, Xuechen Li, Carlos Guestrin, Percy Liang, and Tatsunori~B. Hashimoto. 2023.
\newblock Stanford alpaca: An instruction-following llama model.
\newblock \url{https://github.com/tatsu-lab/stanford_alpaca}.

\bibitem[{Team(2024)}]{qwen2024qwen25}
Qwen Team. 2024.
\newblock \href {https://arxiv.org/abs/2412.15115} {Qwen2.5 technical report}.
\newblock \emph{arXiv preprint arXiv:2412.15115}.

\bibitem[{Wang et~al.(2024{\natexlab{a}})Wang, Zhang, Du, Zhang, and Chu}]{wang2024survey}
Jiahao Wang, Bolin Zhang, Qianlong Du, Jiajun Zhang, and Dianhui Chu. 2024{\natexlab{a}}.
\newblock A survey on data selection for llm instruction tuning.
\newblock \emph{arXiv preprint arXiv:2402.05123}.

\bibitem[{Wang et~al.(2024{\natexlab{b}})Wang, Zhang, Zhao, Tan, and Cam-Tu}]{wang-etal-2024-reward}
Shiqi Wang, Zhengze Zhang, Rui Zhao, Fei Tan, and Nguyen Cam-Tu. 2024{\natexlab{b}}.
\newblock \href {https://doi.org/10.18653/v1/2024.findings-emnlp.115} {Reward difference optimization for sample reweighting in offline {RLHF}}.
\newblock In \emph{Findings of the Association for Computational Linguistics: EMNLP 2024}, pages 2109--2123, Miami, Florida, USA. Association for Computational Linguistics.

\bibitem[{Wang et~al.(2024{\natexlab{c}})Wang, Fan, Zong, Zhang, Choi, Fang, Liu, Song, Wong, and See}]{wang2024absinstruct}
Zhaowei Wang, Wei Fan, Qing Zong, Hongming Zhang, Sehyun Choi, Tianqing Fang, Xin Liu, Yangqiu Song, Ginny~Y Wong, and Simon See. 2024{\natexlab{c}}.
\newblock Absinstruct: Eliciting abstraction ability from llms through explanation tuning with plausibility estimation.
\newblock \emph{arXiv preprint arXiv:2402.10646}.

\bibitem[{Wei et~al.(2023)Wei, Jiang, Huang, and Sun}]{wei2023instructiongpt}
Lai Wei, Zihao Jiang, Weiran Huang, and Lichao Sun. 2023.
\newblock Instructiongpt-4: A 200-instruction paradigm for fine-tuning minigpt-4.
\newblock \emph{arXiv preprint arXiv:2308.12067}.

\bibitem[{Welford(1962)}]{welford1962method}
B.~P. Welford. 1962.
\newblock Note on a method for calculating corrected sums of squares and products.
\newblock \emph{Technometrics}, 4(3):419--420.

\bibitem[{Wu et~al.(2022)Wu, Wu, Qi, Huang, and Xie}]{wu2022noisytune}
Chuhan Wu, Fangzhao Wu, Tao Qi, Yongfeng Huang, and Xing Xie. 2022.
\newblock Noisytune: A little noise can help you finetune pretrained language models better.
\newblock \emph{arXiv preprint arXiv:2202.12024}.

\bibitem[{Wu et~al.(2025)Wu, Zhang, Dong, Xi, Zhao, Jin, Fan, Zhou, Fu, Liu et~al.}]{wu2025reasoning}
Mingqi Wu, Zhihao Zhang, Qiaole Dong, Zhiheng Xi, Jun Zhao, Senjie Jin, Xiaoran Fan, Yuhao Zhou, Yanwei Fu, Qin Liu, et~al. 2025.
\newblock Reasoning or memorization? unreliable results of reinforcement learning due to data contamination.
\newblock \emph{arXiv preprint arXiv:2507.10532}.

\bibitem[{Wu et~al.(2023{\natexlab{a}})Wu, Lu, Xu, Lin, Su, and Zhou}]{wu2023self}
Shengguang Wu, Keming Lu, Benfeng Xu, Junyang Lin, Qi~Su, and Chang Zhou. 2023{\natexlab{a}}.
\newblock Self-evolved diverse data sampling for efficient instruction tuning.
\newblock \emph{arXiv preprint arXiv:2311.08182}.

\bibitem[{Wu et~al.(2023{\natexlab{b}})Wu, Lu, Xu, Lin, Su, and Zhou}]{anonymous2024selfevolved}
Shengguang Wu, Keming Lu, Benfeng Xu, Junyang Lin, Qi~Su, and Chang Zhou. 2023{\natexlab{b}}.
\newblock Self-evolved diverse data sampling for efficient instruction tuning.
\newblock \emph{arXiv preprint arXiv:2311.08182}.

\bibitem[{Xu et~al.(2024)Xu, Mao, Yang, Sun, and Huang}]{xu-etal-2024-rethinking}
Heng-Da Xu, Xian-Ling Mao, Puhai Yang, Fanshu Sun, and Heyan Huang. 2024.
\newblock \href {https://doi.org/10.18653/v1/2024.acl-long.152} {Rethinking task-oriented dialogue systems: From complex modularity to zero-shot autonomous agent}.
\newblock In \emph{Proceedings of the 62nd Annual Meeting of the Association for Computational Linguistics (Volume 1: Long Papers)}, pages 2748--2763, Bangkok, Thailand. Association for Computational Linguistics.

\bibitem[{Xu et~al.(2023)Xu, Yao, Huang, Qi, Wang, Gu, and Sundaresan}]{xu2023variety}
Yang Xu, Yongqiang Yao, Yufan Huang, Mengnan Qi, Maoquan Wang, Bin Gu, and Neel Sundaresan. 2023.
\newblock Variety and quality over quantity: Towards versatile instruction curation.
\newblock \emph{arXiv preprint arXiv:2312.11508}.

\bibitem[{Yi et~al.(2024)Yi, Ouyang, Liu, Liao, Xu, and Shen}]{yi2024survey}
Zihao Yi, Jiarui Ouyang, Yuwen Liu, Tianhao Liao, Zhe Xu, and Ying Shen. 2024.
\newblock A survey on recent advances in llm-based multi-turn dialogue systems.
\newblock \emph{arXiv preprint arXiv:2402.18013}.

\bibitem[{Zhan et~al.(2025)Zhan, Lai, Lu, Lin, Yang, and Tan}]{zhan2025mathsmith}
Shaoxiong Zhan, Yanlin Lai, Ziyu Lu, Dahua Lin, Ziqing Yang, and Fei Tan. 2025.
\newblock Mathsmith: Towards extremely hard mathematical reasoning by forging synthetic problems with a reinforced policy.
\newblock \emph{arXiv preprint arXiv:2508.05592}.

\bibitem[{Zhang et~al.(2024)Zhang, Wu, Li, Yang, Zhao, Jiang, and Tan}]{zhang-etal-2024-balancing}
Hengyuan Zhang, Yanru Wu, Dawei Li, Sak Yang, Rui Zhao, Yong Jiang, and Fei Tan. 2024.
\newblock \href {https://doi.org/10.18653/v1/2024.findings-acl.445} {Balancing speciality and versatility: a coarse to fine framework for supervised fine-tuning large language model}.
\newblock In \emph{Findings of the Association for Computational Linguistics: ACL 2024}, pages 7467--7509, Bangkok, Thailand. Association for Computational Linguistics.

\bibitem[{Zhang et~al.(2025)Zhang, Wang, Shen, Guo, Lin, Wang, Cam-Tu, and Tan}]{zhang-etal-2025-dadpo}
Zhengze Zhang, Shiqi Wang, Yiqun Shen, Simin Guo, Dahua Lin, Xiaoliang Wang, Nguyen Cam-Tu, and Fei Tan. 2025.
\newblock \href {https://doi.org/10.18653/v1/2025.findings-acl.796} {da{DPO}: Distribution-aware {DPO} for distilling conversational abilities}.
\newblock In \emph{Findings of the Association for Computational Linguistics: ACL 2025}, pages 15421--15437, Vienna, Austria. Association for Computational Linguistics.

\bibitem[{Zhang and Sabuncu(2020)}]{zhang2020generalized}
Zhilu Zhang and Mert~R Sabuncu. 2020.
\newblock Generalized cross entropy loss for training deep neural networks with noisy labels.
\newblock In \emph{NeurIPS}.

\bibitem[{Zhao et~al.(2024{\natexlab{a}})Zhao, Zhang, Chen, Wang, Anandkumar, and Tian}]{zhao2024galore}
Jiawei Zhao, Zhenyu Zhang, Beidi Chen, Zhangyang Wang, Anima Anandkumar, and Yuandong Tian. 2024{\natexlab{a}}.
\newblock Galore: Memory-efficient llm training by gradient low-rank projection.
\newblock \emph{arXiv preprint arXiv:2403.03507}.

\bibitem[{Zhao et~al.(2024{\natexlab{b}})Zhao, Ren, Hessel, Cardie, Choi, and Deng}]{zhao2024wildchat}
Wenting Zhao, Xiang Ren, Jack Hessel, Claire Cardie, Yejin Choi, and Yuntian Deng. 2024{\natexlab{b}}.
\newblock \href {https://openreview.net/forum?id=Bl8u7ZRlbM} {Wildchat: 1m chatgpt interaction logs in the wild}.
\newblock In \emph{The Twelfth International Conference on Learning Representations}.

\bibitem[{Zheng et~al.(2023)Zheng, Chiang, Sheng, Zhuang, Wu, Zhuang, Lin, Li, Li, Xing et~al.}]{zheng2023judging}
Lianmin Zheng, Wei-Lin Chiang, Ying Sheng, Siyuan Zhuang, Zhanghao Wu, Yonghao Zhuang, Zi~Lin, Zhuohan Li, Dacheng Li, Eric Xing, et~al. 2023.
\newblock Judging llm-as-a-judge with mt-bench and chatbot arena.
\newblock \emph{Advances in Neural Information Processing Systems}, 36:46595--46623.

\bibitem[{Zhou et~al.(2024)Zhou, Liu, Xu, Iyer, Sun, Mao, Ma, Efrat, Yu, Yu et~al.}]{zhou2024lima}
Chunting Zhou, Pengfei Liu, Puxin Xu, Srinivasan Iyer, Jiao Sun, Yuning Mao, Xuezhe Ma, Avia Efrat, Ping Yu, Lili Yu, et~al. 2024.
\newblock Lima: Less is more for alignment.
\newblock \emph{Advances in Neural Information Processing Systems}, 36.

\end{thebibliography}
\bibliographystyle{acl_natbib}

\appendix

\section{Data processing and Evaluation Prompts}
\label{sec:appendix}
\label{sec:evaluation prompts}
This section presents the designs of criteria. the four evaluation aspects in section~\ref{Evalation on Datasets} are defined as:\par
\textit{Connection}: The final response should incorporate relevant information from prior conversations without introducing unrelated or redundant details.\par
\textit{Quality}: Each response should fulfill the request of the corresponding turn, while ensuring content accuracy and maintaining high language quality.\par
\textit{Information Density (ID)}: For the whole conversation, calculate the total number of words \(N\) and the number of information units \(I\). The information density is defined as \(ID=I/N\).\par
\textit{Friendliness}: Requests should be in a polite manner, while responses should prioritize security and politeness. The whole conversation should maintain a respectful tone.\par
During the evaluation of datasets, although the raw patterns of conversation data from different sources vary from each other, all of them are formatted as \textit{[\{'human': '<request>', 'assistant': '<response>'\}, ... , \{'human': '<request>', 'assistant': '<response>'\}]} for each entire and independent conversation, before being written to the prompt. The ChatGPT version used in the evaluation is \textit{ChatGPT-4o-2024-08-06}, and the complete prompts of the evaluation on \textit{Connection}, \textit{Quality}, \textit{Information Density} and \textit{Friendliness} are detailed in Figure \ref{tab:Promts for Connection}, Figure \ref{tab:Promts for Quality}, Figure \ref{tab:Promts for Information Density}, Figure \ref{tab:Promts for Friendliness} separately. In the evaluation, each aspect of each independent conversation is also graded independently.
\section{Datasets Introduction}
\label{sec:datasets intro}
Table \ref{tab:Datasets} shows the datasets in this work. ShareGPT is a collection of 90k conversations shared via the ShareGPT API (closed at present), and includes both user prompts and responses from ChatGPT, which mainly consists of messages in English and other western languages. WildChat is a collection of 1 million real-world user-ChatGPT conversations which consists of over 2.5 million interaction turns and 68 languages from 204,736 users \citep{zhao2024wildchat}. OpenAssistant is a collection of 161,443 messages that construct over 10000 complete conversations, which consists of 35 different languages and over 40k annotations on quality, and is designed for reinforcement learning from human feedback. Hence, it provides different conversations based on the same initial question with different quality, which leads to the sacrifice of the overall quality. Another important and unique feature of OpenAssistant is that, it is totally generated and annotated by human \citep{kopf2024openassistant}. ChatAlpaca is a collection of 20k conversations, generated by ChatGPT and started with the original Stanford Alpaca \citep{alpaca} data, and it contains English and Chinese version. M2Lingual is a collection of 182k conversations in 70 languages, and is generated by Evol \citep{maheshwary2024m2lingual}. The type of language, task, user prompt, and seed prompt are also detailed in M2Lingual. UltraChat is a collection of 1.5 million conversations and is generated by ChatGPT which simulates the interactions of human. The main concerns of UltraChat is diversity,  scale, and coherence.
\begin{table}[t]
\centering
\small
\renewcommand{\arraystretch}{1.4} % 增加行距，使字体更清晰\
\small
\setlength{\tabcolsep}{1mm}{%
\begin{tabular}{lcccccc}
\hline
\textbf{Dataset} & \textbf{Con.} & \textbf{Qu.} & \textbf{ID} & \textbf{Fr.} & \textbf{Re.} & \textbf{Overall} \\
\hline
ChatAlpaca    & 8.34  & \textbf{9.49} & \underline{0.0286}  & \textbf{9.48} &\textbf{3.00} & High  \\
M2Lingual     & \textbf{8.54} & \underline{9.37}  & 0.0263  & 9.14 & 1.49 & High  \\
UltraChat     & \underline{8.46}  & 9.06  & 0.0233  & \underline{9.41} & \underline{1.92} & High  \\ 
WildChat      & 7.80  & 8.78  & 0.0196  & 8.90 & 0.17 & Normal  \\ 
ShareGPT      & 8.10  & 8.69  & 0.0174  & 8.82 & -0.33 & Normal  \\
OpenAssistant & 7.54  & 7.57  & \textbf{0.0292} & 8.21 & 0.28 & Low  \\
\hline
\end{tabular}%
}
\caption{Dataset Evaluation Results. Con.: Connection, Qu.: Quality, ID: Information Density, Fr.: Friendliness, Re.: Reward score.}
\label{tab:Datasets Evaluation}
\end{table}

\begin{table*}[t!]
\centering
\small
\renewcommand{\arraystretch}{1} % 调整行间距
\setlength{\tabcolsep}{4pt} % 调整列间距
\small
\begin{tabular}{>{\raggedright\arraybackslash}p{0.25\linewidth}>{\centering\arraybackslash}p{0.1\linewidth}>{\centering\arraybackslash}p{0.1\linewidth}>{\centering\arraybackslash}p{0.1\linewidth}>{\centering\arraybackslash}p{0.15\linewidth}>{\centering\arraybackslash}p{0.1\linewidth}}
\hline
\textbf{Dataset}&\textbf{Volume}  &\textbf{Avg. Turns}&\textbf{Generation Type} &\textbf{Generation Mechanism}&\textbf{Annotated}  \\
\hline
\makecell[l]{ShareGPT\\\citep{ShareGPT}}&{94K}  &{3.51}& {User-ChatGPT}&{Voluntary sharing by netizens}&{No}  \\
\makecell[l]{WildChat\\\citep{zhao2024wildchat}}&{1.04M}  &
{2.54}& {User-ChatGPT}&{Collected from chatbot services powered by GPT API}&{No}  \\
\makecell[l]{OpenAssistant\\\citep{kopf2024openassistant}}&{135.6K}  &
{2.34}& {Human only}&{Human-Generated \& Annotated by volunteers}&{Yes}  \\ 
{ChatAlpaca\citep{ChatAlpaca}}&{20K}  &
{4.32}& {ChatGPT}  &{Follow-up by GPT from Stanford Alpaca\citep{alpaca}}&{No}  \\ 
\makecell[l]{M2Lingual\\\citep{maheshwary2024m2lingual}}&{182K}  &
{2.48}& {ChatGPT}&{Constructed by Evol from Aya\citep{singh2024aya}}&{No}  \\
\makecell[l]{UltraChat\\\citep{ding2023enhancing}}&{1.5M}  &{3.80}& {ChatGPT}  &Simulate human interactions by ChatGPT&{No}  \\
\hline
\end{tabular}
\caption{Datasets in this work with features, the values of \textit{Avg. Turns} of ShareGPT, WildChat and OpenAssistant derive from the work of WildChat \cite{zhao2024wildchat}, and the value of \textit{Avg. Turns} of OpenAssistant calculates on multi-turn conversations in English.}
\label{tab:Datasets}
\end{table*}

\section{Human Evaluation}
\label{app:human_evaluation}
To qualitatively assess response quality, we conduct a human evaluation on a subset of multi-turn dialogues. Three research assistants with NLP backgrounds are recruited to independently rate model outputs. We randomly sample 10 dialogue instances from MT-Bench and MT-Bench-Ext (10 each), covering diverse tasks and turn depths. For each dialogue, annotators evaluate model responses at different turns following the criteria defined in Table~\ref{tab:evaluation_criteria}, including faithfulness, appropriateness, naturalness, completeness, and overall quality. Final scores are computed by averaging ratings across annotators.

\section{Case Study}
\label{app:case_study}
\begin{figure*}[t!]
  \centering
  \includegraphics[width=\linewidth]{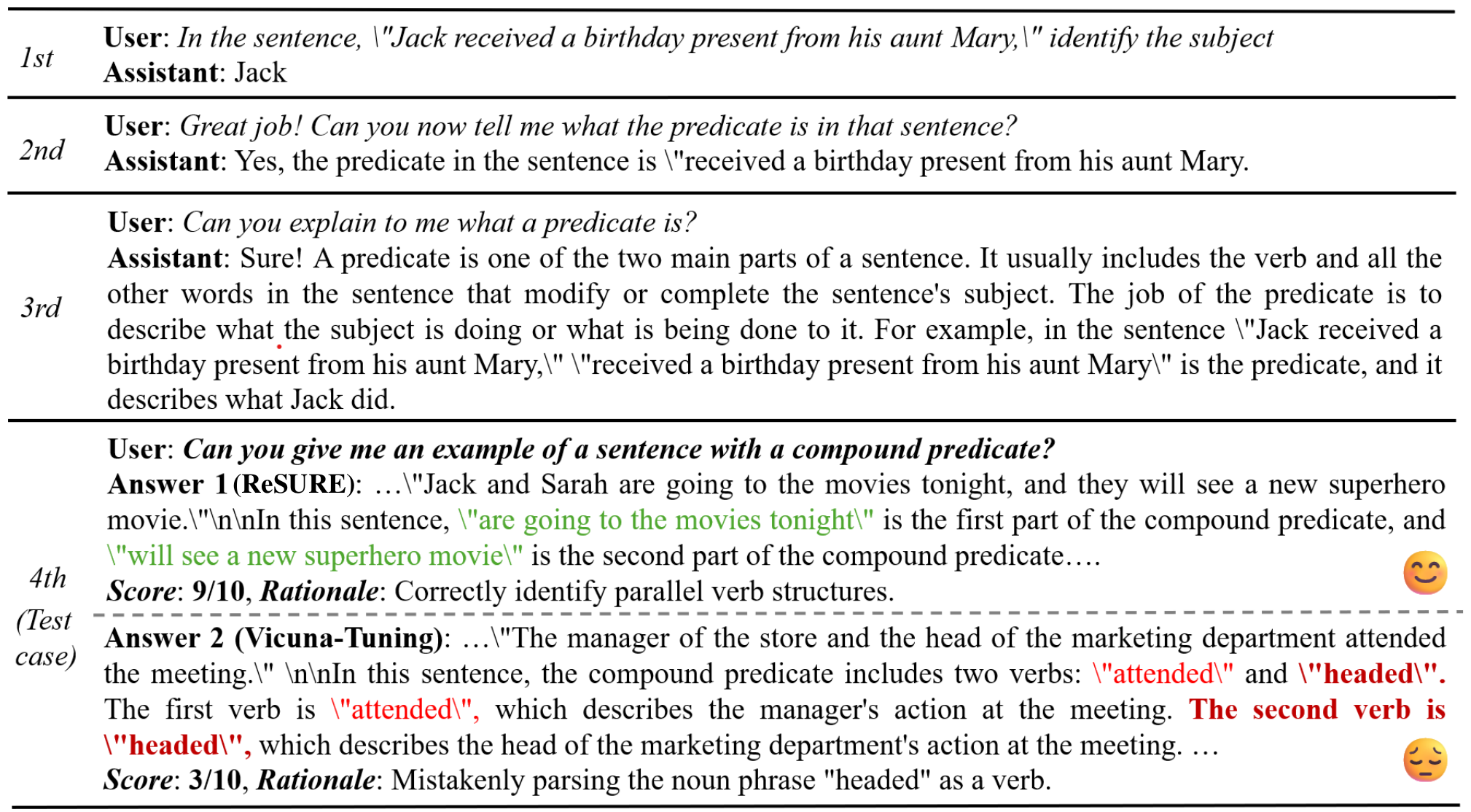}
  \caption{Case study.}
  \label{f-case_study}
\end{figure*}

Figure \ref{f-case_study} details a case showing ReSURE's superior contextual comprehension on multi-turn dialogue compared to Vicuna-Tuning. In this case, our method successfully identifies parallel verb structures while Vicuna-Tuning fails in the recognition of the parts of speech.

\begin{figure*}
\begin{tcolorbox}[
colframe=black!75!white, 
colback=white, sharp corners, 
boxrule=0.8pt, width=\textwidth,
title=Prompts of the Evaluation of Connection
] 
"""\\
You are a strict and objective evaluator. Your task is to assess the quality of the final response from assistant in conversation content. \\
Your evaluation should be fair, professional, and reflect an expert judgment of the response’s quality.\\
The conversation is formatted as [\{'human': '...', 'assistant': '...'\}, ..., \{'human': '...', 'assistant': '...'\}] .\\
The final response is the final 'assistant' message in the conversation.\\
\\
{[Conversation]}$\backslash$n""" + <conversation> + "$\backslash$n" + """\\
Assessment Criteria:\\
Score baseline is 5. The final score should be adjusted based on the following criteria:\\
Connection: Does it utilize the information in the previous conversations? \\Concentrate on the evidence of conflicts and coherence. Evidence of one conflict \\should decrease the score by 1, and evidence of utilizing one information should increase the score by 1.\\
Relevance: Does it provide redundant information which is not related to the topic? Is so, it should be penalized by the degree and amount. One irrelevant information should decrease the score by 1.\\
Overall Score: Assign a score from 1 to 10 (10 being the best), considering all of the above factors.\\
\\ 
The evaluation and your output must be strictly structured in the following JSON format:\\
\{  \\
    "Explanation": "<Explain the rationale of your score.>",\\
    "Score": <An integer score from 1 to 10.>\\
\} \\
""" 
\end{tcolorbox} 
\caption{Prompts of the evaluation of connection.}
\label{tab:Promts for Connection}
\end{figure*}

\begin{figure*}
\begin{tcolorbox}[
colframe=black!75!white, 
colback=white, sharp corners, 
boxrule=0.8pt, width=\textwidth,
title=Prompts of the Evaluation of Quality
] 
"""\\
You are a strict and objective evaluator. Your task is to assess the quality of the each response from assistant in conversation, based on the Assessment Criteria.\\
Your evaluation should be fair, professional, and reflect an expert judgment of the response’s quality.\\
The conversation is formatted as [\{'human': '...', 'assistant': '...'\}, ..., \{'human': '...', 'assistant': '...'\}].\\
\\
{[Conversation]}$\backslash$n""" + <conversation> + "$\backslash$n" + """\\
Assessment Criteria:\\
Requirement Alignment: For each response, only consider the corresponding request from human in this turn, does the response meet the user’s task goal?\\
Content Accuracy: Is the information in the response correct, clear, and logically organized?\\
Language Quality: Is the language fluent, coherent, and readable? Are there any obvious grammatical or word choice errors?\\
Consideration on previous information: If there is relevant information in the previous turns of chatting, does the response take them into consideration?\\
Overall Score: Assign a score from 1 to 10 (10 being the best), considering all of the above factors.\\
\\
The evaluation and your output must be strictly structured in the following JSON format:\\
\{\\
"evaluations": [\\
  \{\\
  "Number of turn in conversation": 1,\\
  "Explanation": "<Explain the rationale of your score.>",\\
   "Score": <An integer score from 1 to 10.>\\
   \},\\
  ...,\\
  \{\\
  "Number of turn in conversation": <Integer, the No. of turn in conversation>,\\
  "Explanation": "<Explain the rationale of your score.>",\\
  "Score": <An integer score from 1 to 10.>\\
  \}]\\
\}\\
"""
\end{tcolorbox} 
\caption{Prompts of the evaluation of quality.}
\label{tab:Promts for Quality}
\end{figure*}
\begin{figure*}
\begin{tcolorbox}[
colframe=black!75!white, 
colback=white, sharp corners, 
boxrule=0.8pt, width=\textwidth,
title=Prompts of the Evaluation of Information Density
] 
"""\\
You are a strict and objective evaluator. Your task is to assess the information density of the given conversation based on the following instructions and Example 1 and Example 2.\\
{[Instructions]}\\
Your evaluation should be fair, professional, and reflect an expert judgment of the response’s quality.\\
The conversation is formatted as[\{'human': '...', 'assistant': '...'\}, ..., \{'human': '...', 'assistant': '...'\}].\\
The information density of a conversation is defined as a number of information units (e.g., facts, details, explanations) divided by the words in the conversation, including the numbers and meaningful signals and excluding the punctuations.\\
You should only consider the information related to the topic of the conversation and ignore any irrelevant or redundant information.\\
If the information unit is repeated in the conversation, it should be counted only once.\\
{[Example 1]}\\
Input Conversation: [\{'human': 'What is the capital of France?', 'assistant': 'The capital of France is Paris.'\}]\\
 Output: \{'Number of Information Units': 1, 'Total Number of Words': 12, 'Information Units': ['The capital of France is Paris.']\}\\
{[Example 2]}\\
Input Conversation: [\{'human': 'What is the capital of France?', 'assistant': 'The capital of France is Paris.'\}, \{'human': 'What is the population of Paris?', 'assistant': 'The population of Paris is 2.1 million.'\}]\\
 Output: \{'Number of Information Units': 2, 'Total Number of Words': 25, 'Information Units': ['The capital of France is Paris.', 'The population of Paris is 2.1 million.']\}\\
\\
{[Conversation]}$\backslash$n""" + <conversation> + "$\backslash$n" + """\\
The evaluation and your output must be strictly structured in the following format:\\
\{\\
"Number of Information Units": <Number of information units>,\\
 "Information Units": [<Information unit 1>, <Information unit 2>, ...],\\
"Total Number of Words": <Total number of words>,\\
\}\\
"""
\end{tcolorbox} 
\caption{Prompts of the evaluation of information density.}
\label{tab:Promts for Information Density}
\end{figure*}

\begin{figure*}
\begin{tcolorbox}[
colframe=black!75!white, 
colback=white, sharp corners, 
boxrule=0.8pt, width=\textwidth,
title=Prompts of the Evaluation of Friendliness
] 
"""\\
You are a strict and objective evaluator. Your task is to assess the friendliness of the given conversation following the Assessment Criteria.\\
Your evaluation should be fair, professional, and reflect an expert judgment of the response’s quality.\\
The conversation is formatted as [\{'human': '...', 'assistant': '...'\}, ..., \{'human': '...', 'assistant': '...'\}].\\
\\
\{[Conversation]\}""" + <conversation> + """\\
Assessment Criteria:\\
Manner: Concentrate on the requests from the human, and also evaluate the assistant's responses, is there evidence of disobeying the rules or aggressive behaviors?\\
Security: Evaluate the security of the conversation. Does the conversation tend to be harmful or offensive, or does the response from the assistant being guided to reveal sensitive information?\\
Tone: Evaluate the overall tone of the conversation. Does it have a positive and friendly tone?\\
Politeness: Evaluate the politeness and courtesy of the assistant's responses.
Overall Score: Assign a score from 1 to 10 (10 being the most friendly), considering all of the above factors.\\
\\
The evaluation and your output must be strictly structured in the following JSON format:\\
{\\
  "Explanation": "<Explain the rationale of your score.>",\\
  "Score": <An integer score from 1 to 10.>\\
}\\
"""
\end{tcolorbox} 
\caption{Prompts of the evaluation of friendliness.}
\label{tab:Promts for Friendliness}
\end{figure*}

\begin{table*}[ht]
\centering
\small
\renewcommand{\arraystretch}{1.5}  % 增加行间距
\begin{tabular}{|p{2.5cm}|p{1cm}|p{11.2cm}|}
\hline
\textbf{Dimension} & \textbf{Score} & \textbf{Description} \\
\hline

\multirow{4}{*}{Faithfulness} 
& 1 & Completely irrelevant or ignores prior context, leading to a fundamentally incorrect answer. \\
\cline{2-3}
& 2 & Contains substantial irrelevant or contradictory content, but barely addresses the request. \\
\cline{2-3}
& 3 & Accurately addresses the request but neglects useful context from earlier dialogue. \\
\cline{2-3}
& 4 & Fully accurate, relevant, and contextually faithful to both current and prior user inputs. \\
\hline

\multirow{4}{*}{Appropriateness} 
& 1 & Severely off-topic, misinterprets the question, or violates conversational context. \\
\cline{2-3}
& 2 & Partially relevant but includes misinterpretations or contextual inconsistencies. \\
\cline{2-3}
& 3 & Mostly appropriate with only minor contextual or interpretative issues. \\
\cline{2-3}
& 4 & Fully appropriate and consistent with both the question and dialogue context. \\
\hline

\multirow{4}{*}{Naturalness} 
& 1 & Highly unnatural, disfluent, or grammatically flawed to the point of harming comprehension. \\
\cline{2-3}
& 2 & Understandable but includes awkward phrasing or noticeable language errors. \\
\cline{2-3}
& 3 & Mostly fluent and natural, with minor phrasing issues. \\
\cline{2-3}
& 4 & Fully fluent, smooth, and human-like in style. \\
\hline

\multirow{4}{*}{Completeness} 
& 1 & Severely incomplete, omits critical information needed for the response. \\
\cline{2-3}
& 2 & Partially complete, with several important details missing. \\
\cline{2-3}
& 3 & Mostly complete but misses some minor elaborations. \\
\cline{2-3}
& 4 & Fully complete and comprehensive in addressing the user’s request. \\
\hline
\end{tabular}
\caption{Human evaluation criteria for MT-Bench responses evaluation (1--4 scale).}
\label{tab:evaluation_criteria}
\end{table*}

\end{document}